\pdfoutput=1

\documentclass[10pt,twocolumn,letterpaper]{article}

\usepackage[pagenumbers]{cvpr}  %

\usepackage{caption}
\usepackage{amsmath,amssymb}
\usepackage{booktabs}
\usepackage{colortbl}%

\usepackage{multirow}
\usepackage{algorithm}
\usepackage{algpseudocode}
\algnotext{EndIf}\algnotext{EndFor}%
\usepackage{tikz}
\usetikzlibrary{arrows.meta,positioning,fit,calc,decorations.pathreplacing}

\newcommand{\R}{\mathbb{R}}

\newcommand{\cC}{\mathcal{C}}

\newcommand{\cP}{\mathcal{P}}

\newcommand{\cY}{\mathcal{Y}}

\newcommand{\fro}[1]{\|#1\|_{F}}

\newcommand{\nuc}[1]{\|#1\|_{*}}

\newcommand{\unf}[2]{{#1}_{(#2)}}
\newcommand{\ours}{\mbox{PERSTAT}}

\DeclareMathOperator{\MAD}{MAD}
\DeclareMathOperator*{\argmin}{arg\,min}
\begingroup\catcode`\_=12 \gdef\authoremail{hasegawa.hiroki.tkb_en@u.tsukuba.ac.jp}\endgroup

\definecolor{cvprblue}{rgb}{0.21,0.49,0.74}
\usepackage[pagebackref,breaklinks,colorlinks,allcolors=cvprblue]{hyperref}
\usepackage[capitalize,nameinlink]{cleveref}
\newcommand{\nvNscenes}{6}
\newcommand{\nvNcases}{18}

\newcommand{\nvNmethods}{16}

\newcommand{\nvShotA}{60}
\newcommand{\nvShotB}{12}
\newcommand{\nvShotC}{3}
\newcommand{\nvNcond}{15}
\newcommand{\nvOursWorstKone}{8.18}
\newcommand{\nvOthersWorstMinKone}{10.54}

\newcommand{\nvOursMeanGapKone}{3.78}
\newcommand{\nvScunetMeanGapKone}{4.42}
\newcommand{\nvOursWithinKone}{4}
\newcommand{\nvOthersWithinMaxKone}{7}
\newcommand{\nvOthersWithinMaxNameKone}{FastDVDnet}
\newcommand{\nvOursWorstKthirteen}{7.90}
\newcommand{\nvOthersWorstMinKthirteen}{10.53}

\newcommand{\nvLearnedImpWorstMinKone}{11.19}
\newcommand{\nvNoImpScunetWorstKone}{5.83}
\newcommand{\nvOursSdGapKone}{2.54}
\newcommand{\nvMedianSdGapKone}{1.80}
\newcommand{\nvMedianMeanGapKone}{8.29}
\newcommand{\nvLearnedSdMinKone}{3.51}
\newcommand{\nvLearnedSdMaxKone}{8.34}

\newcommand{\nvOursMeanGapSevFive}{3.95}
\newcommand{\nvFdvMeanGapSevFive}{3.05}

\newcommand{\nvShiftMeanGapSevFive}{3.52}

\newcommand{\nvOursLpipsAll}{0.35}
\newcommand{\nvScunetLpipsAll}{0.34}

\newcommand{\nvOursLossGaussKone}{4.77}

\newcommand{\nvOursLossPoisKone}{3.52}

\newcommand{\nvOursLossMixKone}{2.14}

\newcommand{\nvOursLossIspKone}{6.94}
\newcommand{\nvOursMedianImpKone}{+7.13}
\newcommand{\nvOursLearnedImpKoneMin}{9.14}

\newcommand{\nvCardNIsp}{36}
\newcommand{\nvSigChange}{+0.86}

\newcommand{\nvOursPoisGain}{$-$1.77}
\newcommand{\nvOursMixGain}{+0.18}

\newcommand{\nvFdvImpGain}{+1.17}

\newcommand{\nvFdvMixGain}{+2.71}

\newcommand{\nvOursNoiseGain}{$-$0.45}

\newcommand{\nvOursRainKone}{25.90}
\newcommand{\nvNoneRainKone}{19.82}

\newcommand{\nvOursSnowKone}{20.88}
\newcommand{\nvNoneSnowKone}{16.63}

\newcommand{\nvOursFogKone}{23.53}
\newcommand{\nvNoneFogKone}{14.98}

\newcommand{\nvNweatherKone}{54}
\newcommand{\nvOursWeatherNbetter}{53}
\newcommand{\nvOursWeatherWorst}{$-$1.00}

\newcommand{\nvAioSnowMin}{2.09}
\newcommand{\nvAioSnowMax}{3.03}
\newcommand{\nvAioLossMin}{2.72}
\newcommand{\nvAioLossMax}{3.86}

\newcommand{\nvMsOursMin}{319}
\newcommand{\nvMsOursMax}{441}

\newcommand{\nvExImpOursPsnr}{39.58}

\newcommand{\nvExImpAioMin}{29.77}
\newcommand{\nvExImpAioMax}{30.05}
\newcommand{\nvExImpLearnedMax}{26.31}

\newcommand{\npNlearned}{14}
\newcommand{\npOursAgreeNoise}{78.51}
\newcommand{\npOthersAgreeNoiseMax}{73.87}
\newcommand{\npOthersAgreeNoiseMaxName}{SCUNet}
\newcommand{\npOursAgreeGauss}{66.48}

\newcommand{\npOursAgreeImp}{90.54}
\newcommand{\npNoneAgreeImp}{51.82}

\newcommand{\npMedianAgreeImp}{78.97}
\newcommand{\npOursMapNoise}{61.35}
\newcommand{\npOthersMapNoiseMax}{50.86}
\newcommand{\npOthersMapNoiseMaxName}{the median filter}
\newcommand{\npOursMapGauss}{28.37}
\newcommand{\npNoneMapGauss}{32.58}

\newcommand{\npOursMapImp}{94.41}

\newcommand{\npMedianMapImp}{72.58}

\newcommand{\npOursAgreeWorst}{66.48}
\newcommand{\npOthersAgreeWorstMax}{64.07}
\newcommand{\npOursMapWorst}{28.37}
\newcommand{\npBestMapWorst}{43.28}
\newcommand{\npBestMapWorstName}{MoCE-IR}
\newcommand{\npOursMapWorstRank}{12}
\newcommand{\npNranked}{15}
\newcommand{\npLearnedBelowNoneImp}{9}

\newcommand{\npLearnedPsnrUpAgreeDownImp}{8}
\newcommand{\npNonePsnrImp}{14.99}
\newcommand{\npNonePsnrGauss}{16.05}
\newcommand{\npUpDownPsnrImpMin}{16.46}
\newcommand{\npUpDownPsnrImpMax}{23.54}

\newcommand{\npDrunetAgreeImp}{46.13}
\newcommand{\npFfdnetAgreeImp}{47.43}

\newcommand{\npLearnedAgreeGaussMin}{73.04}
\newcommand{\npLearnedAgreeGaussMax}{83.67}
\newcommand{\npLearnedMapGaussMin}{40.72}
\newcommand{\npLearnedMapGaussMax}{56.37}
\newcommand{\npOursNearGauss}{40.46}
\newcommand{\npNoneNearGauss}{26.23}

\newcommand{\npMedianMapGauss}{28.88}
\newcommand{\npMedianNearGauss}{31.35}
\newcommand{\npNeverAgreeCleanMin}{88.56}
\newcommand{\npNeverAgreeCleanMax}{91.71}

\newcommand{\npCleanFrames}{539}

\newcommand{\npCleanKept}{504}

\newcommand{\npMacroFoneClean}{0.26}

\newcommand{\npOursPsnrGaussR}{23.50}

\newcommand{\npNframes}{539}

\newcommand{\spOursRain}{25.90}
\newcommand{\spOursRainGain}{+6.07}

\newcommand{\spOursRainNbetter}{18}

\newcommand{\spOursSnow}{20.88}
\newcommand{\spOursSnowGain}{+4.24}

\newcommand{\spOursSnowNbetter}{17}
\newcommand{\spOursSnowWorstCase}{$-$1.00}

\newcommand{\spOursFog}{23.53}
\newcommand{\spOursFogGain}{+8.55}

\newcommand{\spOursFogNbetter}{18}

\newcommand{\spDenoiserWeatherMaxAbs}{1.05}
\newcommand{\spDenoiserWeatherSecondAbs}{0.35}

\newcommand{\spAdaIRSnowGain}{+2.09}
\newcommand{\spAdaIRFogGain}{$-$3.14}
\newcommand{\spPromptIRRainGain}{$-$2.72}
\newcommand{\spPromptIRSnowGain}{+3.03}
\newcommand{\spPromptIRFogGain}{$-$3.86}
\newcommand{\spMoCEIRRainGain}{$-$3.34}
\newcommand{\spMoCEIRSnowGain}{+2.49}

\newcommand{\spViwsRainGain}{+0.19}
\newcommand{\spViwsSnowGain}{+1.58}
\newcommand{\spViwsFogGain}{+0.58}
\newcommand{\spToldRainGain}{+0.30}

\newcommand{\spInstrSnowGain}{+0.19}

\newcommand{\spOursKthirteenKept}{54}
\newcommand{\spNweatherCases}{54}
\newcommand{\spNgateWeather}{108}
\newcommand{\spNcases}{18}
\newcommand{\spDustNoneSA}{24.30}
\newcommand{\spDustNoneSB}{16.85}
\newcommand{\spDustNoneSC}{14.07}
\newcommand{\spDustViwsKoneGain}{+5.66}
\newcommand{\spDustInstrKoneGain}{+2.97}
\newcommand{\spDustOursKoneGain}{+2.09}
\newcommand{\spDustAdaIRKoneGain}{+1.86}
\newcommand{\spDustPromptIRKoneGain}{$-$1.37}
\newcommand{\spDustMoCEIRKoneGain}{$-$0.93}
\newcommand{\spDustDenoiserMaxAbsKone}{0.65}
\newcommand{\spDustViwsKthirteenGain}{+5.08}

\newcommand{\spDustOursGainSA}{+3.02}
\newcommand{\spDustOursGainSB}{+3.16}
\newcommand{\spDustOursGainSC}{+0.07}
\newcommand{\spDustOursNbetter}{16}
\newcommand{\spDustOursNworse}{2}
\newcommand{\spDustOursWorst}{$-$4.45}
\newcommand{\spDustNcases}{18}

\newcommand{\spEtwoBestRainAgree}{72.73}

\newcommand{\spEtwoBestRainMap}{71.16}

\newcommand{\spEtwoBestSnowAgree}{69.14}

\newcommand{\spEtwoBestSnowMap}{56.96}

\newcommand{\spEtwoBestFogAgree}{70.81}
\newcommand{\spEtwoNoneFogAgree}{35.87}

\newcommand{\spEtwoBestFogMap}{85.86}
\newcommand{\spEtwoNoneFogMap}{70.49}
\newcommand{\spEtwoOursMaxAbs}{0.06}
\newcommand{\spEtwoOursKeptMin}{97.40}
\newcommand{\spLotoImpScunetWorst}{5.83}
\newcommand{\spLotoImpOursWorstKone}{8.18}
\newcommand{\spLotoImpOursWorstKthirteen}{7.90}
\newcommand{\spLotoImpOursMeanRankKone}{7th}
\newcommand{\spLotoImpOursMeanRankKthirteen}{3rd}

\newcommand{\spPairWorstScunetLo}{$-$2.84}
\newcommand{\spPairWorstScunetHi}{7.43}

\def\confName{CVPR}
\def\confYear{2027}

\title{\ours{}: Training-Free Statistical Denoising for Vehicle Cameras\\with the Smallest Worst-Case Gap across Noise Types}

\author{Hiroki Hasegawa\\
Graduate School of Science and Technology\\
University of Tsukuba\\
{\tt\small \authoremail}
\and
Hasuka Katano\\
School of Science and Engineering\\
University of Tsukuba\\
{\tt\small s2513549@u.tsukuba.ac.jp}
}

\begin{document}
\twocolumn[{%
\renewcommand\twocolumn[1][]{#1}%
\maketitle
\vspace{-1.6em}
\begin{center}
\captionsetup{type=figure}
\includegraphics[width=\textwidth]{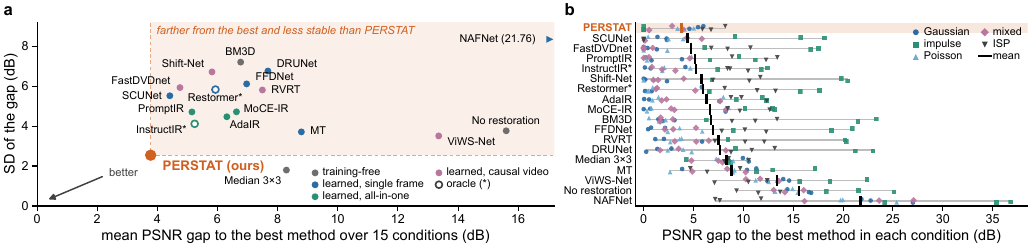}
\captionof{figure}{\textbf{Without training, \ours{} is on average the closest to the best method across five noise types at three severities, and every other method except the median filter is also less stable.}
Each gap is the PSNR gap to the best method in one of the \nvNcond{} conditions (noise type $\times$ severity) of Exp.~1 (Boreas, $k=1$, \ie only the current frame noisy). (a)~Mean and standard deviation (SD) of the gaps of each method. The shaded region is worse than \ours{} in both. (b)~All gaps of each method, with the methods sorted by the mean gap (bar).}
\label{fig:balance}
\end{center}

\vspace{0.6em}
}]
\begin{abstract}
Deep denoisers achieve high accuracy under the noise they were trained on.
However, on a vehicle camera, the noise changes in type and level during a drive.
These denoisers can degrade outside their training noise, so covering more conditions requires more data or retraining.
We propose \ours{}, a training-free statistical optimization method that achieves stable, competitive accuracy across noise conditions rather than the best accuracy under one condition.
\ours{} stacks the latest frames into a tensor and measures the noise of the current frame against the past reference, the noise-free frames just before it.
It then repairs impulse-corrupted color channels from the intact channels of each pixel, shrinks in closed form the noise along the least-variance color direction of the past frames, and applies total variation bounded by the measured noise.
Across 15 noise conditions on Boreas, \ours{} achieves the smallest worst-case PSNR gap to the best method (\nvOursWorstKone{}\,dB), whereas the other \nvNmethods{} methods have gaps of at least \nvOthersWorstMinKone{}\,dB.
Experiments on KITTI show that, in front of frozen driving and detection models, \ours{} preserves the most decisions and detections averaged over Gaussian and impulse noise.
Under Gaussian noise alone, however, it detects fewer objects than most learned methods.
\end{abstract}

\section{Introduction}\label{sec:intro}

Deep denoisers achieve high accuracy under the noise they were trained on.
However, they can degrade under noise distributions and levels outside training.
Gaussian-trained denoisers lose accuracy on real camera noise~\cite{guo2019cbdnet}, overfit to the noise levels seen in training~\cite{mohan2020biasfree} and fail on other noise types~\cite{chen2023maskedtraining}.
Covering more conditions therefore requires more training data or retraining.
Generalizable denoisers train on mixtures of noise~\cite{zhang2023scunet}, all-in-one restorers train on several degradations~\cite{potlapalli2023promptir,cui2025adair}, and some methods keep separate weights for each noise type and level~\cite{zamir2022restormer}.
Generalization across degradations remains open~\cite{yin2025rninet}.

On a vehicle, the camera noise changes during a drive and is hard to predict.
At a tunnel, the light changes abruptly~\cite{bertozzi2011tunnel}, and the noise rises with the camera gain~\cite{wei2020eld}.
Electromagnetic interference and transmission errors can corrupt bands of a frame~\cite{jiang2023glitchhiker}, and faulty pixels and bit errors appear as impulses~\cite{chan2005saltpepper}.
Under such changes, a denoiser that excels on one noise type but fails on another may be less useful than one that stays close to the best on every type.
However, benchmarks that rank methods by their mean score across corruption types~\cite{hendrycks2019benchmarking} can mask a large drop on a single type.
As in worst-group evaluation~\cite{sagawa2020groupdro}, we therefore judge a denoiser by its worst case.

We take a statistical optimization approach without deep learning or pretraining.
Such denoisers need no training data, because they estimate the noise from the noisy image~\cite{donoho1994ideal,liu2013noiselevel} and regularize the image with convex priors such as total variation (TV)~\cite{rudin1992tv}.
However, they must separate texture from noise within the noisy image itself.
On a vehicle, by contrast, the noise-free frames just before the noise starts show nearly the same scene intact.
We call them the \emph{past reference}.
References have restored content before~\cite{ma2011previous,agnolucci2024tape}.
To our knowledge, however, no online denoiser measures the noise of the current frame against one (\cref{sec:related}).

We propose \ours{} (Past-Estimated Reference Subspace for Tensor completion and Adaptive TV), a training-free online denoiser that measures the noise of the current frame against the past reference.
\ours{} stacks the latest 13 frames into a tensor with axes for rows, columns, time and color.
Across frames, a \emph{noise card} compares the current frame with the past reference and estimates the noise level, the impulse type and the outlier pixels without labels.
Within each pixel, \ours{} uses the principal color axes of the past, \ie a PCA of its colors and the color factor of a higher-order SVD~\cite{delathauwer2000hosvd}, to repair the color channels that impulses corrupted.
A nuclear norm then shrinks in closed form only the color component along the least-variance color direction of the past, and TV removes the remaining noise.
The measured statistics, not training, set every step, and a frame without noise goes only to a weather check against the latest frames of the past reference.
We aim not at the best accuracy for each noise type but at stable, competitive accuracy across noise types and severities, that is, the smallest worst-case gap to the best method.

We ask two questions.
\emph{(Q1)} At onsets, the first noisy frames, in Boreas~\cite{burnett2023boreas} clips with five noise types at three severities, how far does \ours{} fall behind the best method, and does the past reference help (\cref{sec:exp-denoise})?
\emph{(Q2)} On noisy KITTI~\cite{geiger2013kitti} drives, does restoration preserve the decisions and detections of frozen models, which a better image does not guarantee (\cref{sec:exp-downstream})?
\Cref{fig:balance} summarizes our answer to Q1, the main result of this paper.%

\noindent\textbf{Contributions.}
\begin{itemize}
\setlength\itemsep{0pt}
\item We propose \ours{}, a statistical optimization denoiser that sets its steps by statistics of the past frames and needs no training data or pretrained weights (\cref{sec:method}).
\item Across five noise types at three severities, \ours{} stays competitive and stable (\cref{sec:exp-denoise}). It achieves the smallest worst-case and mean gaps to the best method, also without a past reference. Although it trails the best learned method under Gaussian, Poisson, mixed and camera-pipeline (ISP) noise, it outperforms all methods under impulse noise and, in PSNR, at synthetic weather onsets.%
\item We design a plug-in protocol with frozen driving and detection models, in which \ours{} preserves the most decisions and detections averaged over both noise types (\cref{sec:protocol,sec:exp-downstream}).
\end{itemize}

\section{Related work}\label{sec:related}

\begin{figure*}[t]
\centering
\input{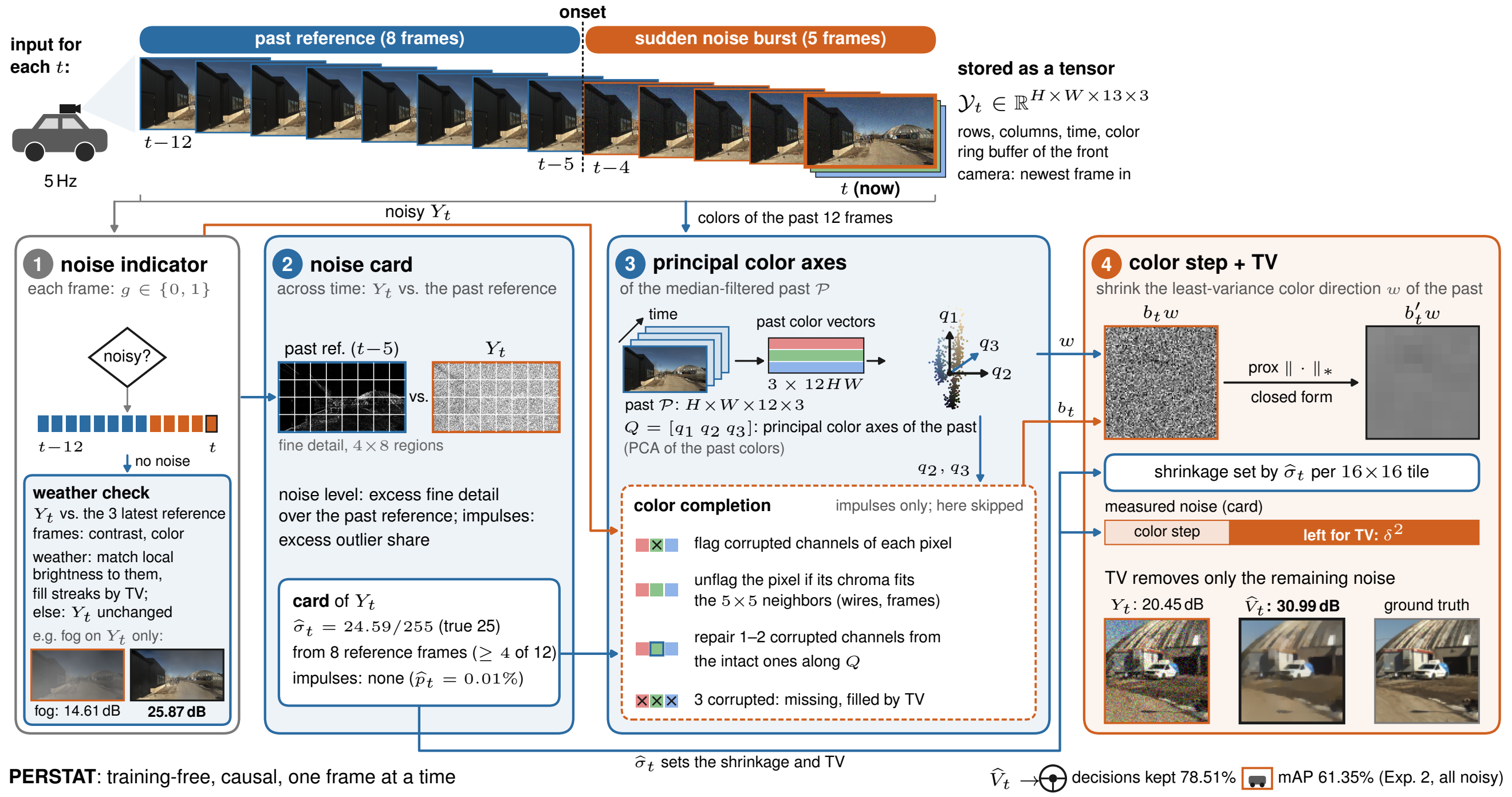}
\caption{\textbf{Without training, \ours{} uses the past reference, the noise-free frames of the same camera, to denoise a suddenly corrupted frame by statistical optimization on the frame tensor and to correct sudden weather on a frame without noise.}
The input holds the 13 latest frames, of which the last $k=5$ are noisy here (Boreas, Gaussian noise, $\sigma=25$). Steps (1)--(4) follow \cref{sec:steps}. The inset of (1) shows fog on $Y_t$ only. Crops are $96\times96$, and PSNR is of the whole frame.}
\label{fig:overview}
\end{figure*}

\noindent\textbf{Generality across noise conditions.}
Generalizable denoisers mask the input~\cite{chen2023maskedtraining} or map unseen noise to Gaussian noise~\cite{ha2026translate}, and all-in-one restorers adapt to trained degradations~\cite{potlapalli2023promptir,cui2025adair}.
However, they all depend on training noise.

\noindent\textbf{Statistical optimization without training.}
Classical denoisers need no training data.
Examples are wavelet shrinkage~\cite{donoho1994ideal} with a noise level estimated from the image~\cite{liu2013noiselevel}, TV~\cite{rudin1992tv}, low-rank tensor models~\cite{zheng2015interpolating}, impulse filters~\cite{chan2005saltpepper} and BM3D~\cite{dabov2007bm3d}.
Zero-shot denoisers instead optimize a network for each image~\cite{ulyanov2018dip}.
However, model-based methods are sensitive to model errors~\cite{shlezinger2023modelbased}, and our card also assumes Gaussian and impulse noise.
Moreover, these methods must separate texture from noise within the noisy image, whereas \ours{} measures the noise against the past reference.

\noindent\textbf{Video restoration with temporal context.}
Video denoisers~\cite{tassano2020fastdvdnet,maggioni2021emvd,liang2022rvrt,ghasemabadi2024turtle} and the training-free V-BM4D~\cite{maggioni2012vbm4d} exploit several frames.
However, they assume noise of the same kind throughout the clip, many use future frames, and none uses its noise-free frames as a reference.

\noindent\textbf{References from the past.}
Reference-based restoration selects the cleanest frames of a video as references for a trained network~\cite{agnolucci2024tape}, low-dose CT restoration borrows structure from a previous scan~\cite{ma2011previous}, and Hindsight~\cite{you2022hindsight} uses past traversals for 3D detection.
Subspace methods estimate a basis from clean or past data~\cite{luo2015targeted,vaswani2018robust}.
In contrast, \ours{} copies no past content and uses the past only as statistics of the noise, the colors and the weather.

\noindent\textbf{Adverse conditions, static and over time.}
Corruption benchmarks~\cite{hendrycks2019benchmarking,dong2023benchmarking} measure accuracy drops of perception models without restoring the image.
Test-time adaptation~\cite{wang2021tent} updates the perception model, which during a drive can cost accuracy or speed~\cite{colomer2023adapt}.
We instead keep the model frozen and restore its input.

\noindent\textbf{Restoration in front of frozen models.}
Placing a denoiser in front of a frozen model is an established design~\cite{liu2018denoisinghighlevel}, but its benefit is not guaranteed.
On real rain and haze, restoration seldom improved frozen detectors~\cite{li2019single,yang2020advancing}, whereas deraining in front of a frozen segmenter outperformed retraining~\cite{porav2019clearly}.
We test denoisers in front of a frozen driving model~\cite{codevilla2019cilrs} and detector on real drives.

\section{\ours{}: statistical optimization}\label{sec:method}
\ours{} restores each new frame in four steps (\cref{fig:overview}, pseudocode in Sec.~S6), each set by statistics of the past frames and solved in closed form or within the measured noise, without training.

\subsection{Setting and the frame tensor}\label{sec:setup}
A camera delivers degraded frames $Y_1,Y_2,\dots$ at any fixed rate, and we use 5\,Hz.
Each frame is an 8-bit RGB image divided by 255, $Y_t\in[0,1]^{H\times W\times3}$ (height $\times$ width $\times$ RGB).
Fixing a pixel $(i,j)$ gives its \emph{color vector} of the three values R, G and B,
\begin{equation}\label{eq:pixel}
y_{ij,t}=\bigl(Y_t(i,j,1),\,Y_t(i,j,2),\,Y_t(i,j,3)\bigr)^{\!\top}\in[0,1]^3 .
\end{equation}
The restorer must output an estimate $\widehat V_t$ of the clean frame $V_t$ as soon as $Y_t$ arrives.
Its input is the frame tensor of the 13 latest frames (rows, columns, time and color),
\begin{equation}\label{eq:tensor}
\cY_t=[\,Y_{t-12},\dots,Y_t\,]\in[0,1]^{H\times W\times13\times3}.
\end{equation}
Windows of 5 to 21 frames change the worst-case gap of Exp.~1 by at most 0.05\,dB (Sec.~S5).

\noindent\textbf{Overall pipeline.}
In Step~1, a binary function $g$ labels each frame of $\cY_t$ noisy or noise-free, and the noise-free past frames form the past reference $\cC_t$.
A noise-free $Y_t$ is returned unchanged, or matched to the latest reference frames if weather degrades it.
For a noisy $Y_t$, Step~2 (noise card) measures its noise against $\cC_t$, Step~3 (color completion) repairs impulse-corrupted color channels, and Step~4 (color step and TV) removes the remaining noise and outputs $\widehat V_t$.
Because the camera moves, no step matches pixels over time.
Steps 1--2 compare statistics across time, and Steps 3--4 work within one frame.

\subsection{Restoring the current frame}\label{sec:steps}
Before the runs, we selected four choices on held-out KITTI drives and set the rest by hand (Sec.~S5).%
\footnote{The constants set by hand are free parameters, which users can tune on validation data of their own camera.}

\noindent\textbf{Step 1: noise indicator and weather check (\cref{fig:overview}(1)).}\label{sec:gate}
To decide whether a frame $Y$ is noisy, Step 1 uses the level $s_\sigma(Y)$ of Gaussian-like noise and the share $s_{\rm imp}(Y)$ of isolated outliers.
$s_\sigma$ uses the Haar diagonal detail of the $2\times2$ block $(i,j)$ of each channel ($p=2i$, $q=2j$), which cancels smooth content but keeps texture and noise,
\begin{equation}\label{eq:haar}
D(Y)_{ij}=\tfrac12\bigl(Y_{p-1,q-1}-Y_{p-1,q}-Y_{p,q-1}+Y_{p,q}\bigr).
\end{equation}
In each $16\times16$ block, $1.4826\operatorname{median}|D(Y)|$ estimates the noise level, where 1.4826 converts the median absolute deviation (MAD) into a Gaussian standard deviation~\cite{donoho1994ideal,rousseeuw1993mad}.
The score $s_\sigma(Y)$ is its 10\% quantile over blocks (the flattest blocks).
$s_{\rm imp}(Y)$ is the share of pixels that differ strongly from their $3\times3$ median and from each of their eight neighbors.
Both scores take the maximum over the color channels.
The noise indicator is the binary function
\begin{equation}\label{eq:gate}
g(Y)=\begin{cases}1 & \text{if } s_\sigma(Y)>\tau_\sigma \text{ or } s_{\rm imp}(Y)>\tau_{\rm imp},\\ 0 & \text{otherwise},\end{cases}
\end{equation}
where 1 means noisy and 0 noise-free, and the past reference $\cC_t=\{s<t:g(Y_s)=0\}$ holds the noise-free past frames.
Even clean frames have nonzero scores, so each threshold $\tau$ is 1.25 times the largest score on clean drives of the same camera.
The factor 1.25 is a hyperparameter, a 25\% margin against labeling clean frames noisy.

A noise-free $Y_t$ goes only through the weather check against $\mathcal{R}_t$, the three latest frames of $\cC_t$.
If its contrast and colorfulness differ from those of $\mathcal{R}_t$, the check matches its local brightness to $\mathcal{R}_t$ and fills thin bright streaks by TV, copying no past pixels (Sec.~S6).
TV sums the color gradients of an image $x$~\cite{rudin1992tv},
\begin{equation}\label{eq:tvdef}
\mathrm{TV}(x)=\sum_{i,j}\fro{\nabla x_{ij}},
\end{equation}
where $x_{ij}\in\R^3$ is the color vector of $x$ at pixel $(i,j)$ and $\nabla x_{ij}=[\,x_{i+1,j}-x_{ij},\ x_{i,j+1}-x_{ij}\,]\in\R^{3\times2}$ its vertical and horizontal differences.
Minimizing TV flattens noise and streaks but tends to keep edges.

\noindent\textbf{Step 2: noise card against the past reference (\cref{fig:overview}(2)).}\label{sec:card}
From the current frame alone, fine texture and noise are hard to distinguish, so the card measures the noise against the past reference.
The fine detail in a region of $Y_t$ is the fine texture of the scene (signal), such as road texture, plus the noise that has just appeared.
The texture already appears in the past frames, so the excess over the past estimates the noise.
Independent noise of level $\sigma$ adds $\sigma^2$ to the expected square of every orthonormal Haar coefficient.
The mean of $D^2$ in the current frame is therefore the energy of the fine texture plus $\sigma^2$, and subtracting that of the past frames estimates $\sigma^2$.
Let $e_r(Y)$ be the mean of $D(Y)^2$ over region $r$ of a $4\times8$ grid, and $\rho(Y)$ the share of outlier pixels (Sec.~S6).
If $|\cC_t|\ge4$ (else Sec.~S6), the card is
\begin{equation}\label{eq:card}
\widehat\sigma_t^2=\max\bigl(0,\operatorname{med}_r[e_r(Y_t)-\bar e_r]\bigr),\ \ \widehat p_t=\rho(Y_t)-\bar\rho ,
\end{equation}
where a bar denotes the mean over the frames of $\cC_t$ and $\operatorname{med}_r$ the median over the 32 regions.
The median is robust to the few regions whose content changed.
Subtracting the past share largely cancels lights, poles and wires, which also pass the outlier test.
The card reports impulses if the outlier share rises by at least 0.5\% ($\widehat p_t\ge0.5\%$), and a Gaussian component if $\widehat\sigma_t\ge3/255$.
Both thresholds are hyperparameters.

\noindent\textbf{Step 3: color completion through the principal color axes (\cref{fig:overview}(3)).}\label{sec:colour}
Color completion repairs impulse-corrupted color channels from the intact channels of the same pixel and the principal color axes of the past.
Let $\cP_t\in\R^{H\times W\times12\times3}$ be the past part of $\cY_t$ after a $3\times3$ median filter of each frame, which keeps impulses out, and $\unf{\cP}{4}\in\R^{3\times12HW}$ the matrix whose columns are its centered color vectors (its mode-4 unfolding~\cite{kolda2009tensor}).
Its principal color axes are the eigenvectors $Q=[q_1\,q_2\,q_3]$ of $\unf{\cP}{4}\unf{\cP}{4}^\top$ (eigenvalues $\Lambda$, decreasing), \ie a PCA of its colors and the color factor of its higher-order SVD~\cite{delathauwer2000hosvd}.
As $q_1$ is close to brightness, $q_2$ and $q_3$ give the chroma $c_a$ of a color vector $y$,
\begin{equation}\label{eq:factor}
\unf{\cP}{4}\unf{\cP}{4}^\top=Q\Lambda Q^\top,\qquad c_a(y)=q_a^\top y,\ \ a=2,3 .
\end{equation}
Because the chroma of a surface usually changes slowly in space, a corrupted channel appears as a jump in $c_a$.
The outlier rule of the card flags each color channel of $Y_t$ separately, and we unflag pixels whose chroma fits the neighborhood, such as wires (Sec.~S6).
Let $\tilde c_a$ and $\MAD_a$ be the median and MAD of $c_a$ over the $5\times5$ neighbors without flagged channels, and $\epsilon=2/255$.
For the flagged channels $B$ of a pixel, we keep its intact channels $B^c$ fixed and choose the values $\xi$ of $B$ so that the repaired chroma is closest to that of the neighborhood.
Chroma axes that vary less in the neighborhood get larger weights $\omega_a=1/(\MAD_a^2+\epsilon^2)$,
\begin{equation}\label{eq:fiber}
y_B\gets\argmin_{\xi\in\R^{|B|}}\sum_{a=2,3}\omega_a\bigl(q_{Ba}^\top \xi-\tilde c_a+q_{B^ca}^\top y_{B^c}\bigr)^2,
\end{equation}
where $q_{Ba}$ and $q_{B^ca}$ hold the entries $B$ and $B^c$ of $q_a$.
A pixel with three flagged channels is missing ($m_{ij}=0$), so we pre-fill it with its $3\times3$ median and leave it to TV.
This tensor completion along the color axis copies no past pixel.
We call the result $b_t$ ($b_t=Y_t$ and all $m_{ij}=1$ without impulses).

\noindent\textbf{Step 4: color step in a subspace of the past, and TV (\cref{fig:overview}(4)).}\label{sec:tv}
Step 4 removes the remaining noise, first along the color direction in which the past varies least and then by TV.
The colors of natural images are strongly correlated and concentrate in few directions~\cite{ruderman1998statistics}, whereas noise that is independent across channels with equal levels is equally strong in every direction.
Let $u_1,u_2,u_3$ be the principal color axes of the unfiltered past frames, by decreasing eigenvalue as in \eqref{eq:factor}.
$U=[u_1\,u_2]$ spans the colors in which the scene varies, and $w=u_3$ is the least-variance color direction.
That $w$ has the least variance holds by construction, whereas that it carries mainly noise is the modeling assumption of \ours{}.
To remove noise along $w$, we write the current frame as the matrix $b_t\in\R^{HW\times3}$ of its color vectors and split it into its parts along $U$ and $w$,
\begin{equation}\label{eq:split}
b_t=b_tUU^\top+o\,w^\top,\qquad o=b_tw\in\R^{HW}.
\end{equation}
We shrink only $o\,w^\top$, tile by tile.
In a $16\times16$ tile $T$, let $f_T\in\R^{256}$ be the values of $o$ minus their tile mean, so that $f_Tw^\top$ is the variation of this part.
The proximal operator of the nuclear norm $\nuc{\cdot}$ (the sum of singular values) shrinks every singular value by $\lambda_t$~\cite{cai2010singular}.
As $f_Tw^\top$ has rank one with singular value $\|f_T\|_2$, it becomes
\begin{equation}\label{eq:shrink}
\operatorname{prox}_{\lambda_t\nuc{\cdot}}\bigl(f_Tw^\top\bigr)=\gamma_Tf_Tw^\top,
\end{equation}
where the shrink factor is ${\gamma_T=(1-\lambda_t/\|f_T\|_2)_+}$ and ${\lambda_t=1.5\,\widehat\sigma_t\sqrt{|T|-1}}$ (without a Gaussian component, see Sec.~S6).
The factor 1.5 is a hyperparameter, a safety margin for shrinking pure noise.
In a tile of pure noise, $\|f_T\|_2\approx\widehat\sigma_t\sqrt{|T|-1}<\lambda_t$ ($|T|-1=255$ degrees of freedom after the mean is removed), so $\gamma_T=0$ and the noise along $w$ is removed.
A tile with a color edge has $\|f_T\|_2\gg\lambda_t$ and keeps most of its variation.
Let $o'$ be $o$ with each tile mean kept and only the variation shrunk by \eqref{eq:shrink}.
As in the subspace norm~\cite{zheng2015interpolating}, singular values thus shrink only within an estimated subspace, and the color step ends in one pass, $b'_t=b_tUU^\top+o'w^\top$.

TV then removes the remaining noise.
It smooths the image but moves the observed pixels $\Omega=\{(i,j):m_{ij}=1\}$ in total no further than the remaining noise $\delta_t$,
\begin{equation}\label{eq:tv}
\widehat V_t\in\argmin_x\ \mathrm{TV}(x)\ \ \text{s.t.}\ \ \|m\odot(x-b'_t)\|_2\le\delta_t ,
\end{equation}
where $\odot$ is the elementwise product, and Chambolle--Pock~\cite{chambolle2011primal} solves \eqref{eq:tv}.
The bound $\delta_t$ estimates the noise that remains in the whole image after the color step.
For a Gaussian component, the noise along $u_1$ and $u_2$ remains in full, whereas along $w$ only the fraction $\bar\gamma_{ij}^2$ remains, with $\bar\gamma_{ij}$ the mean $\gamma_T$ at pixel $(i,j)$ (Sec.~S6), so that
\begin{equation}\label{eq:delta}
\delta_t^2=\widehat\sigma_t^2\textstyle\sum_{(i,j)\in\Omega}\bigl(2+\bar\gamma_{ij}^2\bigr).
\end{equation}
Without a Gaussian component, observed pixels need almost no change, so $\delta_t$ allows only the small error of 8-bit quantization.
TV then mainly fills the missing pixels.
Ordinary TV needs a tuned regularization weight.
Our constraint instead moves the image only as far as the estimated noise (the discrepancy principle~\cite{morozov1966discrepancy}), so no weight is tuned.

\section{Evaluation protocol}\label{sec:protocol}

\begin{table*}[t]
\centering
\caption{\textbf{\ours{} has the smallest worst-case gap to the best method across five noise types and three severities.} Exp.~1, Boreas. Left, mean PSNR (dB) over \nvNcases{} cases per type. Right, gaps to the best method over the \nvNcond{} conditions (type $\times$ severity), namely the worst-case and mean PSNR gaps at $k=1$, the worst-case PSNR gap at $k=13$, the worst-case SSIM and LPIPS gaps at $k=1$, and the mean onset gain $\Delta$. ms is per noisy frame (A100, $^\dagger$CPU). Bold and underline mark the best and second (unrounded). The oracles Restormer$^\ast$ and InstructIR$^\ast$ are not ranked.}
\label{tab:balance}
\footnotesize
\setlength{\tabcolsep}{3.6pt}%
\renewcommand{\arraystretch}{0.93}
\begin{tabular}{@{}l ccccc ccccc c r@{}}
\toprule
 & \multicolumn{5}{c}{PSNR at the onset ($k=1$)} & \multicolumn{3}{c}{PSNR gap (dB)$\downarrow$} & \multicolumn{2}{c}{worst gap$\downarrow$} & & \\
\cmidrule(lr){2-6}\cmidrule(lr){7-9}\cmidrule(lr){10-11}
Method & Gauss. & Impulse & Poisson & Mixed & ISP & worst & mean & worst$_{13}$ & SSIM & LPIPS & $\Delta$ & ms \\
\midrule
No restoration & 15.95 & 15.74 & 15.65 & 13.12 & 15.46 & 25.10 & 15.60 & 25.09 & 0.81 & 1.05 & 0 & 0 \\
Median $3\times3$ & 22.20 & \underline{30.26} & 21.29 & 20.80 & 17.90 & \underline{10.54} & 8.29 & \underline{10.53} & 0.69 & 0.60 & 0 & 4 \\
BM3D$^\dagger$~\cite{dabov2007bm3d} & 29.71 & 20.36 & 28.74 & 25.43 & 15.81 & 23.33 & 6.77 & 23.32 & 0.77 & 0.79 & 0 & 44455 \\
\cmidrule(l{2pt}r{2pt}){1-13}
Restormer$^\ast$~\cite{zamir2022restormer} & 29.62 & 24.08 & 29.56 & 24.94 & 16.05 & 17.63 & 5.93 & 17.63 & 0.68 & 0.59 & 0 & 273 \\
NAFNet~\cite{chen2022nafnet} & 8.01 & 5.27 & 8.48 & 5.25 & 18.09 & 36.82 & 21.76 & 36.82 & 0.99 & 1.38 & 0 & 46 \\
MT~\cite{chen2023maskedtraining} & 23.37 & 23.58 & 23.28 & 19.70 & \underline{20.05} & 17.17 & 8.79 & 17.16 & 0.69 & 0.97 & 0 & 1569 \\
SCUNet~\cite{zhang2023scunet} & 29.87 & 23.04 & 27.88 & 24.10 & \textbf{26.95} & 18.14 & \underline{4.42} & 18.13 & 0.60 & 0.63 & 0 & 117 \\
DRUNet~\cite{zhang2022dpir} & 30.41 & 19.58 & 26.44 & 23.06 & 16.04 & 22.98 & 7.68 & 22.97 & 0.76 & 0.81 & 0 & 38 \\
FFDNet~\cite{zhang2018ffdnet} & 30.77 & 21.14 & 26.21 & 24.08 & 16.87 & 20.93 & 6.97 & 20.92 & 0.72 & 0.83 & 0 & 22 \\
\cmidrule(l{2pt}r{2pt}){1-13}
AdaIR~\cite{cui2025adair} & 29.15 & 26.23 & 26.72 & 23.60 & 16.65 & 12.35 & 6.31 & 12.35 & 0.68 & 0.84 & 0 & 342 \\
PromptIR~\cite{potlapalli2023promptir} & 30.26 & 28.26 & 28.40 & \underline{25.81} & 15.44 & 13.40 & 5.15 & 13.40 & 0.68 & \underline{0.59} & 0 & 303 \\
MoCE-IR~\cite{zamfir2025moceir} & 28.92 & 25.32 & 26.97 & 22.83 & 16.72 & 13.14 & 6.63 & 13.13 & 0.68 & 0.86 & 0 & 292 \\
InstructIR$^\ast$~\cite{conde2024instructir} & 30.21 & 26.85 & 27.30 & 25.58 & 17.78 & 12.41 & 5.24 & 12.40 & 0.56 & 0.59 & 0 & 42 \\
\cmidrule(l{2pt}r{2pt}){1-13}
FastDVDnet~\cite{tassano2020fastdvdnet} & \textbf{31.31} & 23.92 & \underline{29.38} & \textbf{27.22} & 18.32 & 17.53 & 4.75 & 18.65 & \underline{0.57} & 0.67 & \textbf{+1.57} & 59 \\
RVRT~\cite{liang2022rvrt} & 28.86 & 21.07 & 26.82 & 22.55 & 17.16 & 20.33 & 7.49 & 20.53 & 0.70 & 0.83 & $-$0.06 & 2345 \\
Shift-Net~\cite{li2023shiftnet} & \underline{31.04} & 21.68 & \textbf{29.53} & 25.38 & 17.21 & 20.45 & 5.82 & 20.48 & 0.68 & 0.79 & \underline{+0.29} & 1651 \\
ViWS-Net~\cite{yang2023viwsnet} & 18.21 & 18.48 & 17.80 & 15.70 & 16.95 & 22.40 & 13.36 & 22.83 & 0.76 & 0.97 & +0.25 & 109 \\
\midrule
\rowcolor[rgb]{1.0,0.957,0.839}
\textbf{\ours{} (ours)} & 26.54 & \textbf{37.39} & 26.01 & 25.08 & 20.01 & \textbf{8.18} & \textbf{3.78} & \textbf{7.90} & \textbf{0.48} & \textbf{0.45} & $-$0.45 & 334 \\
\bottomrule
\end{tabular}
\end{table*}

\begin{figure*}[t]
\centering
\includegraphics[width=\textwidth]{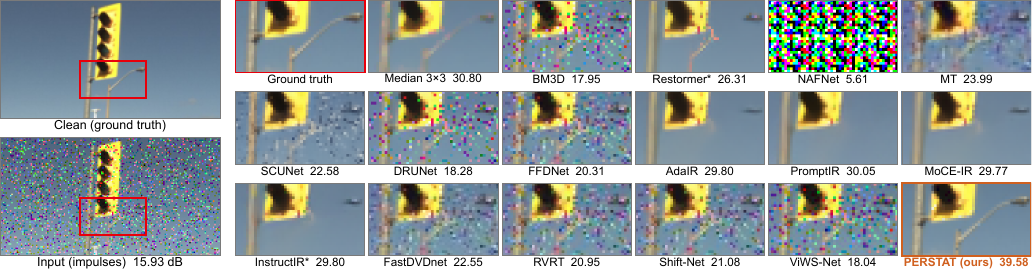}
\caption{\textbf{What is removed and what is kept under impulse noise} (Exp.~1, scene 1 fixed before the run, impulse noise of severity 3 on all 13 frames, $k=13$). Left, clean and noisy crops with the zoomed box. Right, the zoom of every output and the PSNR (dB) of the whole frame. \ours{} removes the impulses and keeps the thin arm of the street lamp (\nvExImpOursPsnr{}\,dB). The all-in-one restorers also remove the impulses but erase the arm (\nvExImpAioMin{}--\nvExImpAioMax{}\,dB). The other learned methods leave impulses or artifacts (${\leq}$\nvExImpLearnedMax{}\,dB).}
\label{fig:onset-qual}
\end{figure*}

\begin{figure*}[t]
\centering
\input{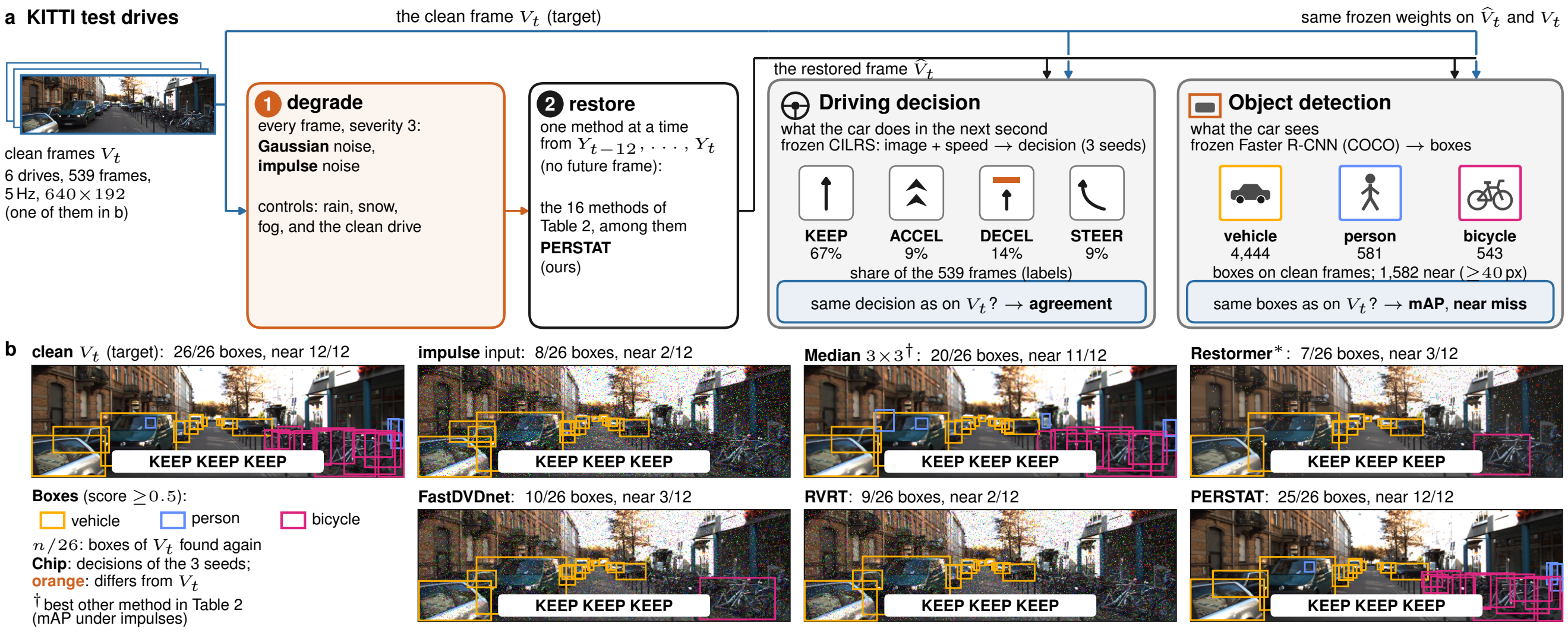}
\caption{\textbf{Exp.~2 measures what the car does and what it sees.} (a)~Every frame is degraded (1) and restored by one method at a time (2). A frozen driving model makes the next driving decision, and a frozen detector finds vehicles, persons and bicycles on $\widehat V_t$ and on the clean $V_t$. A restorer is scored by how many decisions and boxes stay the same. (b)~A test frame chosen before the run by a rule on clean frames, under impulse noise (Fig.~S2 shows Gaussian noise). \ours{} finds 25 of its 26 target boxes, the median filter 20, and the learned methods at most 10.}
\label{fig:exp2-design}
\end{figure*}

\begin{table*}[t]
\centering
\caption{\textbf{\ours{} preserves the most decisions and detections on average over the two noise types.} Exp.~2, \npNframes{} KITTI frames at severity 3, with the metrics of \cref{fig:exp2-design}. Noise is the mean over Gaussian and impulse noise (for mAP, over their pooled frames), and Worst is the lower of the two. ms is per frame (A100). Bold and underline mark the best and second. The oracles ($^\ast$) and no restoration on clean frames are not ranked.}
\label{tab:onset-downstream}
\footnotesize
\setlength{\tabcolsep}{2.9pt}%
\renewcommand{\arraystretch}{0.93}
\begin{tabular}{@{}l ccccc ccccc cc r@{}}
\toprule
 & \multicolumn{5}{c}{Decisions as on the clean frame (\%)} & \multicolumn{5}{c}{Detection mAP@0.5 (\%)} & \multicolumn{2}{c}{Near miss (\%)$\downarrow$} & \\
\cmidrule(lr){2-6}\cmidrule(lr){7-11}\cmidrule(lr){12-13}
Method & Gauss. & Imp. & Noise & Worst & Clean & Gauss. & Imp. & Noise & Worst & Clean & Gauss. & Imp. & ms \\
\midrule
No restoration & 43.78 & 51.82 & 47.80 & 43.78 & 100.00 & 32.58 & 27.38 & 29.80 & 27.38 & 100.00 & 26.23 & 33.38 & 2 \\
Median $3\times3$ & 44.71 & \underline{78.97} & 61.84 & 44.71 & 83.55 & 28.88 & \underline{72.58} & \underline{50.86} & 28.88 & 78.51 & 31.35 & \underline{5.69} & 4 \\
\cmidrule(l{2pt}r{2pt}){1-14}
Restormer$^\ast$~\cite{zamir2022restormer} & 83.12 & 47.74 & 65.43 & 47.74 & 99.63 & 46.83 & 36.69 & 41.70 & 36.69 & 99.91 & 17.26 & 31.92 & 150 \\
NAFNet~\cite{chen2022nafnet} & 43.85 & 42.98 & 43.41 & 42.98 & 85.16 & 0.00 & 0.00 & 0.00 & 0.00 & 81.54 & 100.00 & 100.00 & 42 \\
MT~\cite{chen2023maskedtraining} & 41.99 & 53.18 & 47.59 & 41.99 & 83.49 & 32.30 & 41.66 & 36.92 & 32.30 & 74.54 & 31.92 & 25.09 & 596 \\
SCUNet~\cite{zhang2023scunet} & \textbf{83.67} & 64.07 & \underline{73.87} & \underline{64.07} & 89.61 & 47.20 & 40.80 & 43.99 & \underline{40.80} & 89.51 & 17.19 & 25.28 & 71 \\
DRUNet~\cite{zhang2022dpir} & 76.50 & 46.13 & 61.32 & 46.13 & 95.92 & 49.02 & 25.19 & 37.17 & 25.19 & \underline{98.50} & \underline{16.43} & 37.74 & 23 \\
FFDNet~\cite{zhang2018ffdnet} & 80.27 & 47.43 & 63.85 & 47.43 & 96.17 & 49.35 & 31.37 & 40.29 & 31.37 & 98.04 & 17.57 & 31.10 & 14 \\
\cmidrule(l{2pt}r{2pt}){1-14}
AdaIR~\cite{cui2025adair} & 81.51 & 53.00 & 67.25 & 53.00 & 91.71 & 40.72 & 40.79 & 40.74 & 40.72 & 92.85 & 22.76 & 24.97 & 195 \\
PromptIR~\cite{potlapalli2023promptir} & 81.45 & 62.52 & 71.99 & 62.52 & 88.56 & 41.51 & 40.20 & 40.84 & 40.20 & 94.65 & 21.11 & 21.49 & 171 \\
MoCE-IR~\cite{zamfir2025moceir} & 81.94 & 47.43 & 64.69 & 47.43 & 93.44 & 45.38 & 43.28 & 44.32 & \textbf{43.28} & 97.66 & 18.58 & 23.39 & 172 \\
InstructIR$^\ast$~\cite{conde2024instructir} & 82.68 & 62.71 & 72.70 & 62.71 & 89.86 & 45.31 & 47.15 & 46.22 & 45.31 & 92.87 & 19.15 & 19.53 & 39 \\
\cmidrule(l{2pt}r{2pt}){1-14}
FastDVDnet~\cite{tassano2020fastdvdnet} & 73.04 & 37.29 & 55.16 & 37.29 & 94.81 & 49.56 & 36.93 & 43.22 & 36.93 & 96.82 & 18.65 & 26.80 & 34 \\
RVRT~\cite{liang2022rvrt} & 76.99 & 42.92 & 59.96 & 42.92 & \underline{96.85} & \underline{51.32} & 35.83 & 43.44 & 35.83 & 98.26 & 19.34 & 27.05 & 922 \\
Shift-Net~\cite{li2023shiftnet} & \underline{82.93} & 40.45 & 61.69 & 40.45 & 96.54 & \textbf{56.37} & 33.49 & 44.94 & 33.49 & 97.87 & \textbf{16.18} & 30.40 & 914 \\
ViWS-Net~\cite{yang2023viwsnet} & 43.10 & 46.75 & 44.93 & 43.10 & 94.00 & 29.86 & 32.68 & 31.07 & 29.86 & 90.19 & 28.95 & 28.63 & 100 \\
\midrule
\rowcolor[rgb]{1.0,0.957,0.839}
\textbf{\ours{} (ours)} & 66.48 & \textbf{90.54} & \textbf{78.51} & \textbf{66.48} & \textbf{99.13} & 28.37 & \textbf{94.41} & \textbf{61.35} & 28.37 & \textbf{99.96} & 40.46 & \textbf{3.60} & 331 \\
\bottomrule
\end{tabular}
\end{table*}

\subsection{Compared methods}\label{sec:protocol-methods}
All methods receive the same degraded frames $Y_{t-12},\dots,Y_t$ and no future frame, and each uses as many of them as its official design allows.
Besides no restoration as a baseline, the single-frame methods are a $3\times3$ median filter per channel, Restormer~\cite{zamir2022restormer}, NAFNet~\cite{chen2022nafnet}, MT~\cite{chen2023maskedtraining}, SCUNet~\cite{zhang2023scunet}, DRUNet~\cite{zhang2022dpir}, FFDNet~\cite{zhang2018ffdnet} and BM3D~\cite{dabov2007bm3d}.
The median filter is one of the simplest training-free baselines robust to outliers.
The all-in-one restorers are AdaIR~\cite{cui2025adair}, PromptIR~\cite{potlapalli2023promptir}, MoCE-IR~\cite{zamfir2025moceir} and InstructIR~\cite{conde2024instructir}.
The video methods, run on past frames only, are FastDVDnet~\cite{tassano2020fastdvdnet}, RVRT~\cite{liang2022rvrt}, Shift-Net~\cite{li2023shiftnet} and ViWS-Net~\cite{yang2023viwsnet} (weather).
Restormer$^\ast$ and InstructIR$^\ast$ are \emph{oracles}, given the degradation type.
DRUNet, FFDNet, BM3D, FastDVDnet, RVRT and Shift-Net receive $\widehat\sigma$, the wavelet MAD estimate averaged over the colors (details in Sec.~S6).
All methods use official code and weights in single precision on an NVIDIA A100 GPU.
BM3D (CPU) appears in Exp.~1 only, EMVD~\cite{maggioni2021emvd} lacks public code, and Turtle~\cite{ghasemabadi2024turtle} lacks public denoising weights.
We did not equalize the training data, the CPU for BM3D, or the past-only input of bidirectional methods.

\subsection{Experiment 1: onset denoising}\label{sec:protocol-denoise}
We take \nvNscenes{} scenes from a clear Boreas~\cite{burnett2023boreas} drive (boreas-2021-09-02-11-42), each with the 13 latest camera frames at 5\,Hz ($640\times352$).
We apply five noise types at severities 1, 3 and 5.
Gaussian, impulse and Poisson noise follow ImageNet-C~\cite{hendrycks2019benchmarking}, and the first two also follow KITTI-C~\cite{dong2023benchmarking}.
Mixed noise adds impulses to Gaussian noise rounded to 8 bits.
ISP noise adds shot and read noise in raw space after inverting the camera pipeline~\cite{brooks2019unprocessing}, so it is correlated across pixels and colors (parameters in Sec.~S6).
We did not retune the card, the noise indicator or the weather check, which we had tested on KITTI-C weather.
Each degradation affects the last $k$ of the 13 frames, with $k=1$ at the onset (only the current frame, with a past reference of 12 frames) and $k=13$ as the control.
The degraded current frame is the same in both settings, so the \emph{onset gain}, the PSNR at $k=1$ minus that at $k=13$, isolates the effect of a past reference.
We also give the clean clips to every method.
\Cref{fig:onset-qual} shows one scene under impulse noise.
We report PSNR, SSIM~\cite{wang2004ssim} (Gaussian window, $\sigma=1.5$) and LPIPS~\cite{zhang2018lpips} (AlexNet) against the clean frame $t$ on 8-bit outputs, as means over the \nvNcases{} cases (scenes $\times$ severities) of each noise type and $k$.
The criterion, fixed before the run, is the worst-case gap over the \nvNcond{} conditions (type $\times$ severity) at $k=1$ and $k=13$, where a gap is the mean PSNR deficit to the best non-oracle method.

\subsection{Experiment 2: plug-in restoration}\label{sec:protocol-kitti}
\noindent\textbf{Purpose.}
A restorer helps only if the models behind it decide and detect as on clean frames, which a better image does not guarantee.
Exp.~2 tests two frozen models trained on clean frames, as retraining needs the degradation known in advance and revalidation is costly~\cite{kalra2016driving}.

\noindent\textbf{Data.}
We use the first 100 frames of the six KITTI raw~\cite{geiger2013kitti} test drives (one drive has 39), \npNframes{} frames in all.
We degrade every frame with Gaussian noise, impulse noise, rain, snow and fog from Exp.~1 at severity 3 (one realization, $k=13$).

\noindent\textbf{Models and metrics.}
The driving model follows CILRS~\cite{codevilla2019cilrs}, with an ImageNet-initialized ResNet-34~\cite{he2016resnet} on frame $t$, a speed encoder and four classes instead of continuous controls.
We train it on 35 clean drives with three seeds.
The model has limited accuracy, with a macro-F1 of \npMacroFoneClean{} on clean frames against labels from the ego-motion, where chance is 0.25.
We therefore compare each decision on a restored frame with that of the same model on the clean frame (\emph{agreement}, mean over the seeds), as metamorphic tests of driving models do~\cite{tian2018deeptest,zhang2018deeproad}.
The detector is Faster R-CNN~\cite{ren2015faster} (ResNet-50 FPN, improved torchvision version, official COCO~\cite{lin2014coco} weights), with car, truck and bus merged.
Its target boxes are its own boxes on the clean frame with score ${\ge}0.5$ and height ${\ge}10$\,px.
We report mAP at IoU 0.5 over the three classes and the miss rate of near objects (height ${\ge}40$\,px), where a miss has no box of its class with score ${\ge}0.5$ and IoU ${\ge}0.5$.
We pair the differences to \ours{} by frame and compute 95\% intervals by a bootstrap over 10-frame blocks (2{,}000 resamples, Sec.~S6).

\section{Experiments}\label{sec:experiments}

\subsection{Experiment 1: across noise types and severities}\label{sec:exp-denoise}
\noindent\textbf{Competitive and stable across noise types and severities.}
No method is the best for every noise type, and \ours{} is the best only under impulse noise (\cref{tab:balance}).
However, under the criterion fixed before the run, \ours{} achieves the smallest worst-case gap, \nvOursWorstKone{}\,dB (ISP, severity 3), whereas the other \nvNmethods{} methods have at least \nvOthersWorstMinKone{}\,dB.
Its worst-case SSIM and LPIPS gaps are also the smallest.
Without a past reference ($k=13$), its worst-case gap is also the smallest (\nvOursWorstKthirteen{} \vs ${\geq}$\nvOthersWorstMinKthirteen{}\,dB).
\ours{} consistently has the smallest worst-case gap at every severity, although FastDVDnet, SCUNet and Shift-Net have smaller mean gaps at severity 5 (\nvFdvMeanGapSevFive{}--\nvShiftMeanGapSevFive{} \vs \nvOursMeanGapSevFive{}\,dB).
\Cref{fig:balance}a shows that its gaps are also the smallest on average (\nvOursMeanGapKone{}\,dB, next SCUNet \nvScunetMeanGapKone{}\,dB) and stable (SD \nvOursSdGapKone{}\,dB \vs \nvLearnedSdMinKone{}--\nvLearnedSdMaxKone{}\,dB for the learned methods).
Only the median filter has a lower SD (\nvMedianSdGapKone{}\,dB), but its mean gap is \nvMedianMeanGapKone{}\,dB (each gap in \cref{fig:balance}b and Sec.~S4).
However, \ours{} is within 1\,dB of the best method in only \nvOursWithinKone{} of the \nvNcond{} conditions (\nvOthersWithinMaxNameKone{} in \nvOthersWithinMaxKone{}), and SCUNet has a better mean LPIPS (\nvScunetLpipsAll{} \vs \nvOursLpipsAll{}).

\noindent\textbf{Where it wins and loses.}
Under impulse noise, \ours{} outperforms every method in all \nvNcases{} cases, by \nvOursMedianImpKone{}\,dB over the median filter and by at least \nvOursLearnedImpKoneMin{}\,dB over the learned methods.
Leaving out one noise type at a time indicates that its worst-case lead comes from not failing where learned methods fail.
Without impulse noise, SCUNet has a smaller worst-case gap (\nvNoImpScunetWorstKone{} \vs \nvOursWorstKone{}\,dB), but without any other type \ours{} keeps the smallest (Sec.~S4).
\ours{} keeps thin structures, which the all-in-one restorers erase (\cref{fig:onset-qual}).
Under the other noise types, \ours{} trails the best method, namely FastDVDnet under Gaussian and mixed noise (\nvOursLossGaussKone{} and \nvOursLossMixKone{}\,dB), Shift-Net under Poisson noise (\nvOursLossPoisKone{}\,dB) and SCUNet, trained on camera noise, under ISP noise (\nvOursLossIspKone{}\,dB).
None of these losses exceeds its worst-case gap of \nvOursWorstKone{}\,dB.
The card labels ISP noise as impulses in all \nvCardNIsp{} cases, which suggests that the correlation of ISP noise across pixels and colors violates the assumptions of the card and the color step.
At \nvMsOursMin{}--\nvMsOursMax{}\,ms per noisy frame, \ours{} exceeds the 100\,ms driving budget~\cite{lin2018architectural}.

\noindent\textbf{Does the past reference help?}
Under noise, the past reference does not improve \ours{}.
Its onset gain averages \nvOursNoiseGain{}\,dB (\nvOursPoisGain{} to \nvOursMixGain{}\,dB), and its worst-case gap is smaller without the reference, whereas FastDVDnet (two past frames) gains \nvFdvImpGain{} to \nvFdvMixGain{}\,dB except under ISP noise.
Under Gaussian noise, the reference raises $\widehat\sigma_t$ by \nvSigChange{}/255, so \ours{} smooths more.

\noindent\textbf{Under weather, the past reference helps.}
The noise indicator labels every weather frame and clean clip noise-free ($g=0$), so they go to the weather check.
At the onset, the weather check raises the mean PSNR of rain, snow and fog from \nvNoneRainKone{}, \nvNoneSnowKone{} and \nvNoneFogKone{}\,dB to \nvOursRainKone{}, \nvOursSnowKone{} and \nvOursFogKone{}\,dB, above every compared method, and improves \nvOursWeatherNbetter{} of the \nvNweatherKone{} cases (worst \nvOursWeatherWorst{}\,dB).
SSIM rises under fog but falls slightly under rain and snow.
\ours{} also keeps all six clean clips unchanged, unlike every other restorer.
All-in-one restorers raise the PSNR of snow by \nvAioSnowMin{}--\nvAioSnowMax{}\,dB but lower that of rain and fog by \nvAioLossMin{}--\nvAioLossMax{}\,dB (Sec.~S1), and under sand dust ViWS-Net gains more (Sec.~S2).

\subsection{Experiment 2: decisions and detection}\label{sec:exp-downstream}
\noindent\textbf{Noise.}
Averaged over the two noise types, \ours{} preserves the most decisions (\npOursAgreeNoise{} \vs \npOthersAgreeNoiseMax{}\%, \npOthersAgreeNoiseMaxName{}) and detections (mAP \npOursMapNoise{}\% \vs \npOthersMapNoiseMax{}\% for \npOthersMapNoiseMaxName{}, \cref{tab:onset-downstream}), with paired 95\% intervals excluding zero against every method.
Under the worse noise type, \ours{} keeps a similar share of decisions to SCUNet (\npOursAgreeWorst{} \vs \npOthersAgreeWorstMax{}\%, interval \spPairWorstScunetLo{} to \spPairWorstScunetHi{}), but its mAP ranks \npOursMapWorstRank{}th of \npNranked{} (\npOursMapWorst{} \vs \npBestMapWorst{}\%, \npBestMapWorstName{}).

\noindent\textbf{Methods fail on different noise types.}
Under impulse noise, \npLearnedBelowNoneImp{} of the \npNlearned{} learned methods keep fewer decisions than no restoration (\npNoneAgreeImp{}\%), including the Gaussian-trained DRUNet and FFDNet (\npDrunetAgreeImp{} and \npFfdnetAgreeImp{}\%).
In contrast, \ours{} keeps \npOursAgreeImp{}\% of the decisions with an mAP of \npOursMapImp{}\%, and it is the best for each of the three seeds (median filter \npMedianAgreeImp{} and \npMedianMapImp{}\%).
Under Gaussian noise, however, every learned method except MT, NAFNet and ViWS-Net outperforms \ours{} (agreement \npLearnedAgreeGaussMin{}--\npLearnedAgreeGaussMax{} \vs \npOursAgreeGauss{}\%, mAP \npLearnedMapGaussMin{}--\npLearnedMapGaussMax{} \vs \npOursMapGauss{}\%).
\ours{} even has a slightly lower mAP than no restoration and the median filter (\npNoneMapGauss{} and \npMedianMapGauss{} \vs \npOursMapGauss{}\%) and misses more near objects (\npOursNearGauss{} \vs \npNoneNearGauss{} and \npMedianNearGauss{}\%).
Fig.~S2 suggests that TV erases small objects, which reduces false alarms but increases misses, the unsafe direction.

\noindent\textbf{A better image is not always a better input.}
Under impulse noise, \npLearnedPsnrUpAgreeDownImp{} of those \npLearnedBelowNoneImp{} methods raise the PSNR (\npUpDownPsnrImpMin{}--\npUpDownPsnrImpMax{} \vs \npNonePsnrImp{}\,dB) but still lose decisions.
Under Gaussian noise, \ours{} also raises the PSNR (\npNonePsnrGauss{} $\to$ \npOursPsnrGaussR{}\,dB) but loses detections.
SCUNet, AdaIR and \mbox{PromptIR} are trained on mixed noise or several degradations, and under noise they are never worse than no restoration.
However, like every learned method, they alter every clean frame and keep only \npNeverAgreeCleanMin{}--\npNeverAgreeCleanMax{}\% of the decisions on clean frames, whereas \ours{} returns \npCleanKept{} of the \npCleanFrames{} clean frames unchanged.
Under fog, the all-in-one restorers trained on haze keep the most decisions and detections (Sec.~S3).

\section{Limitations, impact and conclusion}\label{sec:conclusion}

\noindent\textbf{Limitations.}
We synthesized the onsets so that clean ground-truth frames exist, but real onsets, such as at a tunnel, also change exposure.
Exp.~1 covers six scenes of one Boreas drive and five synthetic noise types, so we give no intervals or tests.
In Exp.~2, agreement on recorded frames does not measure closed-loop driving performance~\cite{codevilla2018offline}.
The driving model has low accuracy (macro-F1 \npMacroFoneClean{}) and varies across seeds.
The card assumes noise independent across pixels and colors and a past that shows the same scene under the same exposure, and we set most constants by hand (Sec.~S5).
The color step shrinks one of three color directions, so it removes at most a third of the energy of noise that is i.i.d.\ across channels.
TV removes the rest but also smooths fine texture and small, distant objects.
The weather check works only while its reference frames are weather-free (Sec.~S6), and we tested only synthetic weather.

\noindent\textbf{Broader impact.}
A restorer can change correct decisions or, like ours, miss near objects, so it should be validated on decisions and detections.
No restorer here is shown to be safe in traffic.
We use only public datasets under their licenses and identify no one, so no ethics review was required.

\noindent\textbf{Conclusion.}
We proposed \ours{}, which denoises vehicle camera frames by statistical optimization on the frame tensor, without training data, pretrained weights or future frames.
On Boreas, it trailed learned methods under Gaussian, Poisson, mixed and ISP noise, but it consistently had the smallest worst-case gap to the best method, overall and at each severity.
It also outperformed all methods under impulse noise and, in PSNR, at the onset of synthetic weather.
On KITTI, it kept the most frozen-model decisions and detections averaged over the two noise types, although it missed more near objects than no restoration under Gaussian noise.
Our results suggest that stable accuracy across noise conditions may not require learning the noise.

\section*{Acknowledgments}
This work was supported by JST SPRING, Grant Number \mbox{JPMJSP2124}, and JST CREST, Grant Number \mbox{JPMJCR24Q1}, Japan.

\section*{LLM Usage Statement}
The authors take full responsibility for all submitted content, regardless of any LLM or AI-tool use.
We gave the full manuscript to an LLM provider, Anthropic (Claude), to obtain reviewer-style feedback.
For writing and editing text (language, restructuring, drafting prose) and for code and implementation, the highest level of involvement was \emph{Generate}, in which the authors reviewed, edited or approved tool-generated work through one-shot or interactive use.
For related work and citations and for figures, diagrams and visual material, it was \emph{Assist}, which means editing or refining author-produced content.
For experiments (design, execution and analysis of results), the level of involvement was also \emph{Assist}.
No LLM or AI tool was used for the data (synthetic data, dataset creation and annotation) or for the research ideas, hypotheses, claims and conclusions.

{
    \small
    \bibliographystyle{ieeenat_fullname}
    \bibliography{refs_ibis,refs_verified,refs_onset,refs_m4sn,refs_generalization,refs_deeplimits,refs_v5,refs_sudden,refs_contemp,refs_v6}

\begin{thebibliography}{67}
\providecommand{\natexlab}[1]{#1}
\providecommand{\url}[1]{\texttt{#1}}
\expandafter\ifx\csname urlstyle\endcsname\relax
  \providecommand{\doi}[1]{doi: #1}\else
  \providecommand{\doi}{doi: \begingroup \urlstyle{rm}\Url}\fi

\bibitem[Agnolucci et~al.(2024)Agnolucci, Galteri, Bertini, and
  Del~Bimbo]{agnolucci2024tape}
Lorenzo Agnolucci, Leonardo Galteri, Marco Bertini, and Alberto Del~Bimbo.
\newblock Reference-based restoration of digitized analog videotapes.
\newblock In \emph{2024 IEEE/CVF Winter Conference on Applications of Computer
  Vision (WACV)}, pages 1648--1657, 2024.

\bibitem[Bertozzi et~al.(2011)Bertozzi, Broggi, Boccalini, and
  Mazzei]{bertozzi2011tunnel}
Massimo Bertozzi, Alberto Broggi, Gionata Boccalini, and Luca Mazzei.
\newblock Fast vision-based road tunnel detection.
\newblock In \emph{Image Analysis and Processing -- {ICIAP} 2011}, pages
  424--433, Berlin, Heidelberg, 2011. Springer Berlin Heidelberg.

\bibitem[Brooks et~al.(2019)Brooks, Mildenhall, Xue, Chen, Sharlet, and
  Barron]{brooks2019unprocessing}
Tim Brooks, Ben Mildenhall, Tianfan Xue, Jiawen Chen, Dillon Sharlet, and
  Jonathan~T. Barron.
\newblock Unprocessing images for learned raw denoising.
\newblock In \emph{Proceedings of the IEEE/CVF Conference on Computer Vision
  and Pattern Recognition (CVPR)}, pages 11028--11037, 2019.

\bibitem[Burnett et~al.(2023)Burnett, Yoon, Wu, Li, Zhang, Lu, Qian, Tseng,
  Lambert, Leung, Schoellig, and Barfoot]{burnett2023boreas}
Keenan Burnett, David~J. Yoon, Yuchen Wu, Andrew~Z. Li, Haowei Zhang, Shichen
  Lu, Jingxing Qian, Wei-Kang Tseng, Andrew Lambert, Keith Y.~K. Leung,
  Angela~P. Schoellig, and Timothy~D. Barfoot.
\newblock Boreas: A multi-season autonomous driving dataset.
\newblock \emph{The International Journal of Robotics Research}, 42\penalty0
  (1--2):\penalty0 33--42, 2023.

\bibitem[Cai et~al.(2010)Cai, Cand{\`e}s, and Shen]{cai2010singular}
Jian-Feng Cai, Emmanuel~J. Cand{\`e}s, and Zuowei Shen.
\newblock A singular value thresholding algorithm for matrix completion.
\newblock \emph{SIAM Journal on Optimization}, 20\penalty0 (4):\penalty0
  1956--1982, 2010.

\bibitem[Chambolle and Pock(2011)]{chambolle2011primal}
Antonin Chambolle and Thomas Pock.
\newblock A first-order primal-dual algorithm for convex problems with
  applications to imaging.
\newblock \emph{Journal of Mathematical Imaging and Vision}, 40\penalty0
  (1):\penalty0 120--145, 2011.

\bibitem[Chan et~al.(2005)Chan, Ho, and Nikolova]{chan2005saltpepper}
Raymond~H. Chan, Chung-Wa Ho, and Mila Nikolova.
\newblock Salt-and-pepper noise removal by median-type noise detectors and
  detail-preserving regularization.
\newblock \emph{IEEE Transactions on Image Processing}, 14\penalty0
  (10):\penalty0 1479--1485, 2005.

\bibitem[Chen et~al.(2023)Chen, Gu, Liu, Magid, Dong, Wang, Pfister, and
  Zhu]{chen2023maskedtraining}
Haoyu Chen, Jinjin Gu, Yihao Liu, Salma~Abdel Magid, Chao Dong, Qiong Wang,
  Hanspeter Pfister, and Lei Zhu.
\newblock Masked image training for generalizable deep image denoising.
\newblock In \emph{Proceedings of the IEEE/CVF Conference on Computer Vision
  and Pattern Recognition (CVPR)}, pages 1692--1703, 2023.

\bibitem[Chen et~al.(2022)Chen, Chu, Zhang, and Sun]{chen2022nafnet}
Liangyu Chen, Xiaojie Chu, Xiangyu Zhang, and Jian Sun.
\newblock Simple baselines for image restoration.
\newblock In \emph{Computer Vision -- {ECCV} 2022}, pages 17--33, Cham, 2022.
  Springer Nature Switzerland.

\bibitem[Codevilla et~al.(2018)Codevilla, L{\'o}pez, Koltun, and
  Dosovitskiy]{codevilla2018offline}
Felipe Codevilla, Antonio~M. L{\'o}pez, Vladlen Koltun, and Alexey Dosovitskiy.
\newblock {On Offline Evaluation of Vision-Based Driving Models}.
\newblock In \emph{{Computer Vision -- ECCV 2018}}, pages 246--262, 2018.

\bibitem[Codevilla et~al.(2019)Codevilla, Santana, L{\'o}pez, and
  Gaidon]{codevilla2019cilrs}
Felipe Codevilla, Eder Santana, Antonio~M. L{\'o}pez, and Adrien Gaidon.
\newblock Exploring the limitations of behavior cloning for autonomous driving.
\newblock In \emph{2019 {IEEE/CVF} International Conference on Computer Vision
  ({ICCV})}, pages 9328--9337, 2019.

\bibitem[Colomer et~al.(2023)Colomer, Dovesi, Panagiotakopoulos, Carvalho,
  H{\"a}renstam-Nielsen, Azizpour, Kjellstr{\"o}m, Cremers, and
  Poggi]{colomer2023adapt}
Marc~Botet Colomer, Pier~Luigi Dovesi, Theodoros Panagiotakopoulos,
  Joao~Frederico Carvalho, Linus H{\"a}renstam-Nielsen, Hossein Azizpour,
  Hedvig Kjellstr{\"o}m, Daniel Cremers, and Matteo Poggi.
\newblock To adapt or not to adapt? {Real-Time} adaptation for semantic
  segmentation.
\newblock In \emph{Proceedings of the IEEE/CVF International Conference on
  Computer Vision (ICCV)}, pages 16502--16513, 2023.

\bibitem[Conde et~al.(2024)Conde, Geigle, and Timofte]{conde2024instructir}
Marcos~V. Conde, Gregor Geigle, and Radu Timofte.
\newblock {InstructIR}: High-quality image restoration following human
  instructions.
\newblock In \emph{Computer Vision -- ECCV 2024}, pages 1--21. Springer, 2024.

\bibitem[Cui et~al.(2025)Cui, Zamir, Khan, Knoll, Shah, and Khan]{cui2025adair}
Yuning Cui, Syed~Waqas Zamir, Salman Khan, Alois Knoll, Mubarak Shah, and
  Fahad~Shahbaz Khan.
\newblock {AdaIR}: Adaptive all-in-one image restoration via frequency mining
  and modulation.
\newblock In \emph{International Conference on Learning Representations
  (ICLR)}, 2025.

\bibitem[Dabov et~al.(2007)Dabov, Foi, Katkovnik, and
  Egiazarian]{dabov2007bm3d}
Kostadin Dabov, Alessandro Foi, Vladimir Katkovnik, and Karen Egiazarian.
\newblock Image denoising by sparse 3-{D} transform-domain collaborative
  filtering.
\newblock \emph{IEEE Transactions on Image Processing}, 16\penalty0
  (8):\penalty0 2080--2095, 2007.

\bibitem[De~Lathauwer et~al.(2000)De~Lathauwer, De~Moor, and
  Vandewalle]{delathauwer2000hosvd}
Lieven De~Lathauwer, Bart De~Moor, and Joos Vandewalle.
\newblock A multilinear singular value decomposition.
\newblock \emph{SIAM Journal on Matrix Analysis and Applications}, 21\penalty0
  (4):\penalty0 1253--1278, 2000.

\bibitem[Dong et~al.(2023)Dong, Kang, Zhang, Zhu, Wang, Yang, Su, Wei, and
  Zhu]{dong2023benchmarking}
Yinpeng Dong, Caixin Kang, Jinlai Zhang, Zijian Zhu, Yikai Wang, Xiao Yang,
  Hang Su, Xingxing Wei, and Jun Zhu.
\newblock Benchmarking robustness of {3D} object detection to common
  corruptions in autonomous driving.
\newblock In \emph{2023 IEEE/CVF Conference on Computer Vision and Pattern
  Recognition (CVPR)}, pages 1022--1032, 2023.

\bibitem[Donoho and Johnstone(1994)]{donoho1994ideal}
David~L. Donoho and Iain~M. Johnstone.
\newblock Ideal spatial adaptation by wavelet shrinkage.
\newblock \emph{Biometrika}, 81\penalty0 (3):\penalty0 425--455, 1994.

\bibitem[Geiger et~al.(2013)Geiger, Lenz, Stiller, and
  Urtasun]{geiger2013kitti}
Andreas Geiger, Philip Lenz, Christoph Stiller, and Raquel Urtasun.
\newblock Vision meets robotics: The {KITTI} dataset.
\newblock \emph{The International Journal of Robotics Research}, 32\penalty0
  (11):\penalty0 1231--1237, 2013.

\bibitem[Ghasemabadi et~al.(2024)Ghasemabadi, Janjua, Salameh, and
  Niu]{ghasemabadi2024turtle}
Amirhosein Ghasemabadi, Muhammad~Kamran Janjua, Mohammad Salameh, and Di Niu.
\newblock Learning truncated causal history model for video restoration.
\newblock In \emph{Advances in Neural Information Processing Systems}, pages
  27584--27615. Curran Associates, Inc., 2024.

\bibitem[Guo et~al.(2019)Guo, Yan, Zhang, Zuo, and Zhang]{guo2019cbdnet}
Shi Guo, Zifei Yan, Kai Zhang, Wangmeng Zuo, and Lei Zhang.
\newblock Toward convolutional blind denoising of real photographs.
\newblock In \emph{Proceedings of the IEEE/CVF Conference on Computer Vision
  and Pattern Recognition (CVPR)}, pages 1712--1722, 2019.

\bibitem[Ha et~al.(2026)Ha, Ryou, Seo, and Han]{ha2026translate}
Inju Ha, Donghun Ryou, Seonguk Seo, and Bohyung Han.
\newblock Learning to translate noise for robust image denoising.
\newblock In \emph{Proceedings of the IEEE/CVF Conference on Computer Vision
  and Pattern Recognition - Findings Track (CVPRF)}, pages 5034--5043, 2026.

\bibitem[He et~al.(2016)He, Zhang, Ren, and Sun]{he2016resnet}
Kaiming He, Xiangyu Zhang, Shaoqing Ren, and Jian Sun.
\newblock Deep residual learning for image recognition.
\newblock In \emph{2016 {IEEE} Conference on Computer Vision and Pattern
  Recognition ({CVPR})}, pages 770--778, 2016.

\bibitem[Hendrycks and Dietterich(2019)]{hendrycks2019benchmarking}
Dan Hendrycks and Thomas Dietterich.
\newblock Benchmarking neural network robustness to common corruptions and
  perturbations.
\newblock In \emph{International Conference on Learning Representations
  ({ICLR})}, 2019.

\bibitem[Jiang et~al.(2023)Jiang, Ji, Yan, Xie, Lou, and
  Xu]{jiang2023glitchhiker}
Qinhong Jiang, Xiaoyu Ji, Chen Yan, Zhixin Xie, Haina Lou, and Wenyuan Xu.
\newblock {GlitchHiker}: Uncovering vulnerabilities of image signal
  transmission with {IEMI}.
\newblock In \emph{32nd USENIX Security Symposium (USENIX Security 23)}, pages
  7249--7266, Anaheim, CA, 2023. USENIX Association.

\bibitem[Kalra and Paddock(2016)]{kalra2016driving}
Nidhi Kalra and Susan~M. Paddock.
\newblock {Driving to Safety: How Many Miles of Driving Would It Take to
  Demonstrate Autonomous Vehicle Reliability?}
\newblock \emph{{Transportation Research Part A: Policy and Practice}},
  94:\penalty0 182--193, 2016.

\bibitem[Kolda and Bader(2009)]{kolda2009tensor}
Tamara~G. Kolda and Brett~W. Bader.
\newblock Tensor decompositions and applications.
\newblock \emph{SIAM Review}, 51\penalty0 (3):\penalty0 455--500, 2009.

\bibitem[Li et~al.(2023)Li, Shi, Zhang, Cheung, See, Wang, Qin, and
  Li]{li2023shiftnet}
Dasong Li, Xiaoyu Shi, Yi Zhang, Ka~Chun Cheung, Simon See, Xiaogang Wang,
  Hongwei Qin, and Hongsheng Li.
\newblock A simple baseline for video restoration with grouped spatial-temporal
  shift.
\newblock In \emph{2023 IEEE/CVF Conference on Computer Vision and Pattern
  Recognition (CVPR)}, pages 9822--9832, 2023.

\bibitem[Li et~al.(2019)Li, Araujo, Ren, Wang, Tokuda, Hirata~Junior,
  Cesar-Junior, Zhang, Guo, and Cao]{li2019single}
Siyuan Li, Iago~Breno Araujo, Wenqi Ren, Zhangyang Wang, Eric~K. Tokuda,
  Roberto Hirata~Junior, Roberto Cesar-Junior, Jiawan Zhang, Xiaojie Guo, and
  Xiaochun Cao.
\newblock Single image deraining: A comprehensive benchmark analysis.
\newblock In \emph{Proceedings of the {IEEE/CVF} Conference on Computer Vision
  and Pattern Recognition ({CVPR})}, pages 3833--3842, 2019.

\bibitem[Liang et~al.(2022)Liang, Fan, Xiang, Ranjan, Ilg, Green, Cao, Zhang,
  Timofte, and Van~Gool]{liang2022rvrt}
Jingyun Liang, Yuchen Fan, Xiaoyu Xiang, Rakesh Ranjan, Eddy Ilg, Simon Green,
  Jiezhang Cao, Kai Zhang, Radu Timofte, and Luc Van~Gool.
\newblock Recurrent video restoration transformer with guided deformable
  attention.
\newblock In \emph{Advances in Neural Information Processing Systems}, pages
  378--393. Curran Associates, Inc., 2022.

\bibitem[Lin et~al.(2018)Lin, Zhang, Hsu, Skach, Haque, Tang, and
  Mars]{lin2018architectural}
Shih-Chieh Lin, Yunqi Zhang, Chang-Hong Hsu, Matt Skach, Md~E. Haque, Lingjia
  Tang, and Jason Mars.
\newblock {The Architectural Implications of Autonomous Driving: Constraints
  and Acceleration}.
\newblock In \emph{{Proceedings of the Twenty-Third International Conference on
  Architectural Support for Programming Languages and Operating Systems
  (ASPLOS)}}, pages 751--766, 2018.

\bibitem[Lin et~al.(2014)Lin, Maire, Belongie, Hays, Perona, Ramanan,
  Doll{\'a}r, and Zitnick]{lin2014coco}
Tsung-Yi Lin, Michael Maire, Serge Belongie, James Hays, Pietro Perona, Deva
  Ramanan, Piotr Doll{\'a}r, and C.~Lawrence Zitnick.
\newblock Microsoft {COCO}: Common objects in context.
\newblock In \emph{Computer Vision -- {ECCV} 2014}, pages 740--755, Cham, 2014.
  Springer International Publishing.

\bibitem[Liu et~al.(2018)Liu, Wen, Liu, Wang, and
  Huang]{liu2018denoisinghighlevel}
Ding Liu, Bihan Wen, Xianming Liu, Zhangyang Wang, and Thomas Huang.
\newblock When image denoising meets high-level vision tasks: A deep learning
  approach.
\newblock In \emph{Proceedings of the Twenty-Seventh International Joint
  Conference on Artificial Intelligence ({IJCAI-18})}, pages 842--848.
  International Joint Conferences on Artificial Intelligence Organization,
  2018.

\bibitem[Liu et~al.(2013)Liu, Tanaka, and Okutomi]{liu2013noiselevel}
Xinhao Liu, Masayuki Tanaka, and Masatoshi Okutomi.
\newblock Single-image noise level estimation for blind denoising.
\newblock \emph{IEEE Transactions on Image Processing}, 22\penalty0
  (12):\penalty0 5226--5237, 2013.

\bibitem[Luo et~al.(2015)Luo, Chan, and Nguyen]{luo2015targeted}
Enming Luo, Stanley~H. Chan, and Truong~Q. Nguyen.
\newblock Adaptive image denoising by targeted databases.
\newblock \emph{IEEE Transactions on Image Processing}, 24\penalty0
  (7):\penalty0 2167--2181, 2015.

\bibitem[Ma et~al.(2011)Ma, Huang, Feng, Zhang, Lu, Liang, and
  Chen]{ma2011previous}
Jianhua Ma, Jing Huang, Qianjin Feng, Hua Zhang, Hongbing Lu, Zhengrong Liang,
  and Wufan Chen.
\newblock Low-dose computed tomography image restoration using previous
  normal-dose scan.
\newblock \emph{Medical Physics}, 38\penalty0 (10):\penalty0 5713--5731, 2011.

\bibitem[Maggioni et~al.(2012)Maggioni, Boracchi, Foi, and
  Egiazarian]{maggioni2012vbm4d}
Matteo Maggioni, Giacomo Boracchi, Alessandro Foi, and Karen Egiazarian.
\newblock Video denoising, deblocking, and enhancement through separable 4-{D}
  nonlocal spatiotemporal transforms.
\newblock \emph{IEEE Transactions on Image Processing}, 21\penalty0
  (9):\penalty0 3952--3966, 2012.

\bibitem[Maggioni et~al.(2021)Maggioni, Huang, Li, Xiao, Fu, and
  Song]{maggioni2021emvd}
Matteo Maggioni, Yibin Huang, Cheng Li, Shuai Xiao, Zhongqian Fu, and Fenglong
  Song.
\newblock Efficient multi-stage video denoising with recurrent spatio-temporal
  fusion.
\newblock In \emph{2021 IEEE/CVF Conference on Computer Vision and Pattern
  Recognition (CVPR)}, pages 3465--3474, 2021.

\bibitem[Mohan et~al.(2020)Mohan, Kadkhodaie, Simoncelli, and
  Fernandez-Granda]{mohan2020biasfree}
Sreyas Mohan, Zahra Kadkhodaie, Eero~P. Simoncelli, and Carlos
  Fernandez-Granda.
\newblock Robust and interpretable blind image denoising via bias-free
  convolutional neural networks.
\newblock In \emph{International Conference on Learning Representations
  (ICLR)}, 2020.

\bibitem[Morozov(1966)]{morozov1966discrepancy}
V.~A. Morozov.
\newblock On the solution of functional equations by the method of
  regularization.
\newblock \emph{Doklady Akademii Nauk SSSR}, 167\penalty0 (3):\penalty0
  510--512, 1966.

\bibitem[Porav et~al.(2019)Porav, Bruls, and Newman]{porav2019clearly}
Horia Porav, Tom Bruls, and Paul Newman.
\newblock {I Can See Clearly Now: Image Restoration via De-Raining}.
\newblock In \emph{{Proceedings of the IEEE International Conference on
  Robotics and Automation (ICRA)}}, pages 7087--7093, 2019.

\bibitem[Potlapalli et~al.(2023)Potlapalli, Zamir, Khan, and
  Khan]{potlapalli2023promptir}
Vaishnav Potlapalli, Syed~Waqas Zamir, Salman Khan, and Fahad~Shahbaz Khan.
\newblock {PromptIR}: Prompting for all-in-one image restoration.
\newblock In \emph{Advances in Neural Information Processing Systems}, pages
  71275--71293, 2023.

\bibitem[Ren et~al.(2015)Ren, He, Girshick, and Sun]{ren2015faster}
Shaoqing Ren, Kaiming He, Ross Girshick, and Jian Sun.
\newblock Faster {R-CNN}: Towards real-time object detection with region
  proposal networks.
\newblock In \emph{Advances in Neural Information Processing Systems}. Curran
  Associates, Inc., 2015.

\bibitem[Rousseeuw and Croux(1993)]{rousseeuw1993mad}
Peter~J. Rousseeuw and Christophe Croux.
\newblock Alternatives to the median absolute deviation.
\newblock \emph{Journal of the American Statistical Association}, 88\penalty0
  (424):\penalty0 1273--1283, 1993.

\bibitem[Ruderman et~al.(1998)Ruderman, Cronin, and
  Chiao]{ruderman1998statistics}
Daniel~L. Ruderman, Thomas~W. Cronin, and Chuan-Chin Chiao.
\newblock Statistics of cone responses to natural images: Implications for
  visual coding.
\newblock \emph{Journal of the Optical Society of America A}, 15\penalty0
  (8):\penalty0 2036--2045, 1998.

\bibitem[Rudin et~al.(1992)Rudin, Osher, and Fatemi]{rudin1992tv}
Leonid~I. Rudin, Stanley Osher, and Emad Fatemi.
\newblock Nonlinear total variation based noise removal algorithms.
\newblock \emph{Physica D: Nonlinear Phenomena}, 60\penalty0 (1--4):\penalty0
  259--268, 1992.

\bibitem[Sagawa et~al.(2020)Sagawa, Koh, Hashimoto, and
  Liang]{sagawa2020groupdro}
Shiori Sagawa, Pang~Wei Koh, Tatsunori~B. Hashimoto, and Percy Liang.
\newblock Distributionally robust neural networks for group shifts: On the
  importance of regularization for worst-case generalization.
\newblock In \emph{International Conference on Learning Representations
  (ICLR)}, 2020.

\bibitem[Shlezinger et~al.(2023)Shlezinger, Whang, Eldar, and
  Dimakis]{shlezinger2023modelbased}
Nir Shlezinger, Jay Whang, Yonina~C. Eldar, and Alexandros~G. Dimakis.
\newblock Model-based deep learning.
\newblock \emph{Proceedings of the {IEEE}}, 111\penalty0 (5):\penalty0
  465--499, 2023.

\bibitem[Tassano et~al.(2020)Tassano, Delon, and Veit]{tassano2020fastdvdnet}
Matias Tassano, Julie Delon, and Thomas Veit.
\newblock {FastDVDnet}: Towards real-time deep video denoising without flow
  estimation.
\newblock In \emph{Proc. IEEE/CVF Conf. Computer Vision and Pattern Recognition
  (CVPR)}, pages 1351--1360, 2020.

\bibitem[Tian et~al.(2018)Tian, Pei, Jana, and Ray]{tian2018deeptest}
Yuchi Tian, Kexin Pei, Suman Jana, and Baishakhi Ray.
\newblock {DeepTest}: Automated testing of deep-neural-network-driven
  autonomous cars.
\newblock In \emph{Proceedings of the 40th International Conference on Software
  Engineering (ICSE)}, pages 303--314, 2018.

\bibitem[Ulyanov et~al.(2018)Ulyanov, Vedaldi, and Lempitsky]{ulyanov2018dip}
Dmitry Ulyanov, Andrea Vedaldi, and Victor Lempitsky.
\newblock Deep image prior.
\newblock In \emph{2018 {IEEE/CVF} Conference on Computer Vision and Pattern
  Recognition ({CVPR})}, pages 9446--9454, 2018.

\bibitem[Vaswani et~al.(2018)Vaswani, Bouwmans, Javed, and
  Narayanamurthy]{vaswani2018robust}
Namrata Vaswani, Thierry Bouwmans, Sajid Javed, and Praneeth Narayanamurthy.
\newblock Robust subspace learning: Robust {PCA}, robust subspace tracking, and
  robust subspace recovery.
\newblock \emph{IEEE Signal Processing Magazine}, 35\penalty0 (4):\penalty0
  32--55, 2018.

\bibitem[Wang et~al.(2021)Wang, Shelhamer, Liu, Olshausen, and
  Darrell]{wang2021tent}
Dequan Wang, Evan Shelhamer, Shaoteng Liu, Bruno Olshausen, and Trevor Darrell.
\newblock Tent: Fully test-time adaptation by entropy minimization.
\newblock In \emph{International Conference on Learning Representations}, 2021.

\bibitem[Wang et~al.(2004)Wang, Bovik, Sheikh, and Simoncelli]{wang2004ssim}
Zhou Wang, Alan~C. Bovik, Hamid~R. Sheikh, and Eero~P. Simoncelli.
\newblock Image quality assessment: from error visibility to structural
  similarity.
\newblock \emph{IEEE Transactions on Image Processing}, 13\penalty0
  (4):\penalty0 600--612, 2004.

\bibitem[Wei et~al.(2020)Wei, Fu, Yang, and Huang]{wei2020eld}
Kaixuan Wei, Ying Fu, Jiaolong Yang, and Hua Huang.
\newblock A physics-based noise formation model for extreme low-light raw
  denoising.
\newblock In \emph{2020 IEEE/CVF Conference on Computer Vision and Pattern
  Recognition (CVPR)}, pages 2755--2764, 2020.

\bibitem[Yang et~al.(2020)Yang, Yuan, Ren, Liu, Scheirer, Wang,
  et~al.]{yang2020advancing}
Wenhan Yang, Ye Yuan, Wenqi Ren, Jiaying Liu, Walter~J. Scheirer, Zhangyang
  Wang, et~al.
\newblock Advancing image understanding in poor visibility environments: A
  collective benchmark study.
\newblock \emph{{IEEE} Transactions on Image Processing}, 29:\penalty0
  5737--5752, 2020.

\bibitem[Yang et~al.(2023)Yang, Aviles-Rivero, Fu, Liu, Wang, and
  Zhu]{yang2023viwsnet}
Yijun Yang, Angelica~I. Aviles-Rivero, Huazhu Fu, Ye Liu, Weiming Wang, and Lei
  Zhu.
\newblock Video adverse-weather-component suppression network via weather
  messenger and adversarial backpropagation.
\newblock In \emph{2023 IEEE/CVF International Conference on Computer Vision
  (ICCV)}, pages 13154--13164, 2023.

\bibitem[Yin et~al.(2025)Yin, Wang, Lin, Ran, and Zheng]{yin2025rninet}
Zhengwei Yin, Hongjun Wang, Guixu Lin, Weihang Ran, and Yinqiang Zheng.
\newblock Random is all you need: Random noise injection on feature statistics
  for generalizable deep image denoising.
\newblock In \emph{International Conference on Learning Representations
  (ICLR)}, pages 51885--51908, 2025.

\bibitem[You et~al.(2022)You, Luo, Chen, Chen, Chao, Sun, Hariharan, Campbell,
  and Weinberger]{you2022hindsight}
Yurong You, Katie~Z. Luo, Xiangyu Chen, Junan Chen, Wei-Lun Chao, Wen Sun,
  Bharath Hariharan, Mark Campbell, and Kilian~Q. Weinberger.
\newblock Hindsight is 20/20: Leveraging past traversals to aid {3D}
  perception.
\newblock In \emph{International Conference on Learning Representations}, 2022.

\bibitem[Zamfir et~al.(2025)Zamfir, Wu, Mehta, Tan, Paudel, Zhang, and
  Timofte]{zamfir2025moceir}
Eduard Zamfir, Zongwei Wu, Nancy Mehta, Yuedong Tan, Danda~Pani Paudel, Yulun
  Zhang, and Radu Timofte.
\newblock Complexity experts are task-discriminative learners for any image
  restoration.
\newblock In \emph{Proceedings of the IEEE/CVF Conference on Computer Vision
  and Pattern Recognition (CVPR)}, pages 12753--12763, 2025.

\bibitem[Zamir et~al.(2022)Zamir, Arora, Khan, Hayat, Khan, and
  Yang]{zamir2022restormer}
Syed~Waqas Zamir, Aditya Arora, Salman Khan, Munawar Hayat, Fahad~Shahbaz Khan,
  and Ming-Hsuan Yang.
\newblock {Restormer}: Efficient transformer for high-resolution image
  restoration.
\newblock In \emph{2022 {IEEE/CVF} Conference on Computer Vision and Pattern
  Recognition ({CVPR})}, pages 5718--5729, 2022.

\bibitem[Zhang et~al.(2018{\natexlab{a}})Zhang, Zuo, and
  Zhang]{zhang2018ffdnet}
Kai Zhang, Wangmeng Zuo, and Lei Zhang.
\newblock {FFDNet}: Toward a fast and flexible solution for {CNN}-based image
  denoising.
\newblock \emph{IEEE Transactions on Image Processing}, 27\penalty0
  (9):\penalty0 4608--4622, 2018{\natexlab{a}}.

\bibitem[Zhang et~al.(2022)Zhang, Li, Zuo, Zhang, Van~Gool, and
  Timofte]{zhang2022dpir}
Kai Zhang, Yawei Li, Wangmeng Zuo, Lei Zhang, Luc Van~Gool, and Radu Timofte.
\newblock Plug-and-play image restoration with deep denoiser prior.
\newblock \emph{IEEE Transactions on Pattern Analysis and Machine
  Intelligence}, 44\penalty0 (10):\penalty0 6360--6376, 2022.

\bibitem[Zhang et~al.(2023)Zhang, Li, Liang, Cao, Zhang, Tang, Fan, Timofte,
  and Van~Gool]{zhang2023scunet}
Kai Zhang, Yawei Li, Jingyun Liang, Jiezhang Cao, Yulun Zhang, Hao Tang,
  Deng-Ping Fan, Radu Timofte, and Luc Van~Gool.
\newblock Practical blind image denoising via {Swin-Conv-UNet} and data
  synthesis.
\newblock \emph{Machine Intelligence Research}, 20\penalty0 (6):\penalty0
  822--836, 2023.

\bibitem[Zhang et~al.(2018{\natexlab{b}})Zhang, Zhang, Zhang, Liu, and
  Khurshid]{zhang2018deeproad}
Mengshi Zhang, Yuqun Zhang, Lingming Zhang, Cong Liu, and Sarfraz Khurshid.
\newblock {DeepRoad}: {GAN}-based metamorphic testing and input validation
  framework for autonomous driving systems.
\newblock In \emph{Proceedings of the 33rd ACM/IEEE International Conference on
  Automated Software Engineering (ASE)}, pages 132--142, 2018{\natexlab{b}}.

\bibitem[Zhang et~al.(2018{\natexlab{c}})Zhang, Isola, Efros, Shechtman, and
  Wang]{zhang2018lpips}
Richard Zhang, Phillip Isola, Alexei~A. Efros, Eli Shechtman, and Oliver Wang.
\newblock The unreasonable effectiveness of deep features as a perceptual
  metric.
\newblock In \emph{2018 {IEEE/CVF} Conference on Computer Vision and Pattern
  Recognition ({CVPR})}, pages 586--595, 2018{\natexlab{c}}.

\bibitem[Zheng and Tomioka(2015)]{zheng2015interpolating}
Qinqing Zheng and Ryota Tomioka.
\newblock Interpolating convex and non-convex tensor decompositions via the
  subspace norm.
\newblock In \emph{Advances in Neural Information Processing Systems}. Curran
  Associates, Inc., 2015.

\end{thebibliography}
}
\clearpage
\appendix
\renewcommand{\thetable}{S\arabic{table}}\setcounter{table}{0}
\renewcommand{\thefigure}{S\arabic{figure}}\setcounter{figure}{0}
\renewcommand{\thesection}{S\arabic{section}}\setcounter{section}{0}
\renewcommand{\theequation}{S\arabic{equation}}\setcounter{equation}{0}
\renewcommand{\thealgorithm}{S\arabic{algorithm}}\setcounter{algorithm}{0}
\twocolumn[{\centering\Large\bfseries Supplementary Material\par\vspace{1.2em}}]

\section{Rain, snow and fog in Experiment 1}\label{sec:supp-weather}
This section gives the weather controls of Exp.~1 that Sec.~\ref{sec:exp-denoise} summarizes.

\noindent\textbf{Setup.}
The six Boreas scenes, the methods and the two settings are those of Exp.~1 (Sec.~\ref{sec:protocol-denoise}): a degradation is applied to the current frame only ($k=1$, with a past reference of 12 frames) or to all 13 frames ($k=13$).
Rain is the rain layer of KITTI-C, and snow and fog are those of ImageNet-C, at severities 1, 3 and 5; as in KITTI-C, rain and snow are then blended with gray and darkened, and fog is blended with gray; the fog pattern is drawn once per clip, so it is the same in all 13 frames.
The oracle Restormer$^\ast$ uses its deraining weights under rain and its blind weights under snow and fog; the oracle InstructIR$^\ast$ receives one fixed instruction per weather type (``Clear the rain from my picture'', ``Remove the snow from my picture'', ``Remove the fog from my photo''); its weights were not trained for snow.
The clean clips are the six scenes without degradation.
Each value is a mean over the \spNcases{} cases (scenes $\times$ severities) of a weather type.

\noindent\textbf{The weather check of \ours{}} (Sec.~\ref{sec:gate}) acts on frames without noise: its reference is the three latest past frames labeled noise-free ($g=0$), so with all 13 frames weather-degraded it has none that differ from the current frame.

\noindent\textbf{Results.}
With a past reference, \ours{} raises the mean PSNR of rain, snow and fog to \spOursRain{}, \spOursSnow{} and \spOursFog{}\,dB (\spOursRainGain{}, \spOursSnowGain{} and \spOursFogGain{}\,dB over the input), above every compared method, and improves \spOursRainNbetter{}, \spOursSnowNbetter{} and \spOursFogNbetter{} of the \spNcases{} cases (worst case \spOursSnowWorstCase{}\,dB, snow); it returns all six clean clips unchanged.
With all 13 frames degraded (\cref{tab:supp-weather-k13}) it returns all \spOursKthirteenKept{} of the \spNweatherCases{} weather cases unchanged.
The denoisers barely change the weather (\cref{tab:supp-weather-k1}): the mean PSNR of rain, snow or fog moves by at most \spDenoiserWeatherMaxAbs{}\,dB (Shift-Net, fog), all others by at most \spDenoiserWeatherSecondAbs{}\,dB.
The all-in-one restorers, trained on Gaussian noise, rain and haze, raise snow (AdaIR \spAdaIRSnowGain{}, PromptIR \spPromptIRSnowGain{}, MoCE-IR \spMoCEIRSnowGain{}\,dB) but lower rain (\spPromptIRRainGain{} to \spMoCEIRRainGain{}\,dB) and fog (\spAdaIRFogGain{} to \spPromptIRFogGain{}\,dB); the weather restorer ViWS-Net raises snow, fog and rain by \spViwsSnowGain{}, \spViwsFogGain{} and \spViwsRainGain{}\,dB; the oracle Restormer$^\ast$ gains \spToldRainGain{}\,dB under rain with its deraining weights, and the oracle InstructIR$^\ast$ at most \spInstrSnowGain{}\,dB.
The noise indicator of \ours{} labels all \spNgateWeather{} weather cases and the clean clips noise-free; every other restorer changes every weather case and all six clean clips, and behaves at $k=13$ as at $k=1$.

\begin{table*}[t]
\centering
\caption{\textbf{Rain, snow and fog at the onset: the weather check of \ours{} restores the most.} Exp.~1, Boreas, degradation on the current frame only ($k=1$). Mean over \spNcases{} cases per weather type. Clean clips: the number of the 6 clips returned unchanged (kept) and the lowest PSNR (dB) of a changed one. Bold/underline: best/second by the unrounded means; the oracle Restormer$^\ast$ and InstructIR$^\ast$, given the degradation type, are not ranked.}
\label{tab:supp-weather-k1}
\small
\setlength{\tabcolsep}{4.0pt}
\begin{tabular}{@{}l ccc ccc ccc cc@{}}
\toprule
 & \multicolumn{3}{c}{Rain} & \multicolumn{3}{c}{Snow} & \multicolumn{3}{c}{Fog} & \multicolumn{2}{c}{Clean clips} \\
\cmidrule(lr){2-4}\cmidrule(lr){5-7}\cmidrule(lr){8-10}\cmidrule(lr){11-12}
Method & PSNR$\uparrow$ & SSIM$\uparrow$ & LPIPS$\downarrow$ & PSNR$\uparrow$ & SSIM$\uparrow$ & LPIPS$\downarrow$ & PSNR$\uparrow$ & SSIM$\uparrow$ & LPIPS$\downarrow$ & kept & min PSNR \\
\midrule
No restoration & 19.82 & \textbf{0.81} & 0.33 & 16.63 & 0.59 & 0.53 & 14.98 & 0.73 & 0.36 & 6 & -- \\
Median $3\times3$ & 19.68 & 0.77 & 0.38 & 16.74 & 0.60 & 0.54 & 14.96 & 0.69 & 0.43 & 0 & 32.21 \\
BM3D & 19.82 & 0.81 & 0.33 & 16.64 & 0.59 & 0.53 & 14.98 & 0.72 & 0.36 & 0 & 54.34 \\
\cmidrule(l{2pt}r{2pt}){1-12}
Restormer$^\ast$ & 20.12 & 0.87 & 0.17 & 16.63 & 0.59 & 0.53 & 14.98 & 0.73 & 0.36 & 0 & 76.72 \\
NAFNet & 19.53 & 0.77 & 0.44 & 16.49 & 0.60 & 0.61 & 14.92 & 0.68 & 0.50 & 0 & 29.85 \\
MT & 19.47 & 0.76 & 0.42 & 16.74 & \underline{0.65} & 0.55 & 14.80 & 0.65 & 0.63 & 0 & 33.59 \\
SCUNet & 19.80 & 0.79 & 0.36 & 16.59 & 0.60 & 0.53 & 14.94 & 0.69 & 0.47 & 0 & 39.29 \\
DRUNet & 19.82 & 0.81 & 0.34 & 16.63 & 0.59 & 0.53 & 14.98 & 0.72 & 0.36 & 0 & 53.33 \\
FFDNet & 19.82 & 0.81 & 0.33 & 16.64 & 0.59 & 0.53 & 14.98 & 0.72 & 0.36 & 0 & 50.34 \\
\cmidrule(l{2pt}r{2pt}){1-12}
AdaIR & 16.76 & 0.78 & 0.32 & 18.72 & 0.55 & 0.49 & 11.84 & 0.59 & \underline{0.25} & 0 & 23.95 \\
PromptIR & 17.10 & 0.79 & \textbf{0.32} & \underline{19.67} & 0.59 & \underline{0.48} & 11.12 & 0.52 & 0.27 & 0 & 23.78 \\
MoCE-IR & 16.48 & 0.77 & 0.32 & 19.12 & 0.57 & 0.49 & 11.20 & 0.52 & 0.27 & 0 & 24.38 \\
InstructIR$^\ast$ & 20.01 & 0.86 & 0.20 & 16.82 & 0.66 & 0.46 & 13.02 & 0.69 & 0.27 & 0 & 43.36 \\
\cmidrule(l{2pt}r{2pt}){1-12}
FastDVDnet & 19.80 & 0.81 & 0.34 & 16.66 & 0.60 & 0.53 & 14.98 & 0.72 & 0.38 & 0 & 44.84 \\
RVRT & 19.82 & \underline{0.81} & 0.33 & 16.64 & 0.59 & 0.53 & 14.98 & 0.73 & 0.36 & 0 & 54.45 \\
Shift-Net & 19.82 & 0.81 & 0.34 & 16.64 & 0.59 & 0.53 & 13.93 & 0.65 & 0.46 & 0 & 51.71 \\
ViWS-Net & \underline{20.02} & 0.79 & 0.34 & 18.21 & \textbf{0.70} & \textbf{0.38} & \underline{15.56} & \underline{0.78} & 0.26 & 0 & 28.34 \\
\midrule
\rowcolor[rgb]{1.0,0.957,0.839}
\textbf{\ours{} (ours)} & \textbf{25.90} & 0.78 & \underline{0.32} & \textbf{20.88} & 0.58 & 0.51 & \textbf{23.53} & \textbf{0.93} & \textbf{0.10} & 6 & -- \\
\bottomrule
\end{tabular}
\end{table*}

\begin{table*}[t]
\centering
\caption{\textbf{The same with all 13 frames degraded ($k=13$).} Layout as in \cref{tab:supp-weather-k1}, without the clean clips.}
\label{tab:supp-weather-k13}
\small
\setlength{\tabcolsep}{4.2pt}
\begin{tabular}{@{}l ccc ccc ccc@{}}
\toprule
 & \multicolumn{3}{c}{Rain} & \multicolumn{3}{c}{Snow} & \multicolumn{3}{c}{Fog} \\
\cmidrule(lr){2-4}\cmidrule(lr){5-7}\cmidrule(lr){8-10}
Method & PSNR$\uparrow$ & SSIM$\uparrow$ & LPIPS$\downarrow$ & PSNR$\uparrow$ & SSIM$\uparrow$ & LPIPS$\downarrow$ & PSNR$\uparrow$ & SSIM$\uparrow$ & LPIPS$\downarrow$ \\
\midrule
No restoration & \underline{19.82} & \textbf{0.81} & 0.33 & 16.63 & 0.59 & 0.53 & 14.98 & \underline{0.73} & 0.36 \\
Median $3\times3$ & 19.68 & 0.77 & 0.38 & 16.74 & 0.60 & 0.54 & 14.96 & 0.69 & 0.43 \\
BM3D & 19.82 & 0.81 & 0.33 & 16.64 & 0.59 & 0.53 & 14.98 & 0.72 & 0.36 \\
\cmidrule(l{2pt}r{2pt}){1-10}
Restormer$^\ast$ & 20.12 & 0.87 & 0.17 & 16.63 & 0.59 & 0.53 & 14.98 & 0.73 & 0.36 \\
NAFNet & 19.53 & 0.77 & 0.44 & 16.49 & 0.60 & 0.61 & 14.92 & 0.68 & 0.50 \\
MT & 19.47 & 0.76 & 0.42 & 16.74 & \underline{0.65} & 0.55 & 14.80 & 0.65 & 0.63 \\
SCUNet & 19.80 & 0.79 & 0.36 & 16.59 & 0.60 & 0.53 & 14.94 & 0.69 & 0.47 \\
DRUNet & 19.82 & 0.81 & 0.34 & 16.63 & 0.59 & 0.53 & 14.98 & 0.72 & 0.36 \\
FFDNet & 19.82 & 0.81 & 0.33 & 16.64 & 0.59 & 0.53 & 14.98 & 0.72 & 0.36 \\
\cmidrule(l{2pt}r{2pt}){1-10}
AdaIR & 16.76 & 0.78 & 0.32 & 18.72 & 0.55 & 0.49 & 11.84 & 0.59 & \textbf{0.25} \\
PromptIR & 17.10 & 0.79 & \textbf{0.32} & \textbf{19.67} & 0.59 & \underline{0.48} & 11.12 & 0.52 & 0.27 \\
MoCE-IR & 16.48 & 0.77 & \underline{0.32} & \underline{19.12} & 0.57 & 0.49 & 11.20 & 0.52 & 0.27 \\
InstructIR$^\ast$ & 20.01 & 0.86 & 0.20 & 16.82 & 0.66 & 0.46 & 13.02 & 0.69 & 0.27 \\
\cmidrule(l{2pt}r{2pt}){1-10}
FastDVDnet & 19.80 & 0.81 & 0.34 & 16.66 & 0.60 & 0.53 & \underline{14.99} & 0.72 & 0.38 \\
RVRT & 19.82 & 0.81 & 0.33 & 16.64 & 0.59 & 0.53 & 14.98 & 0.73 & 0.36 \\
Shift-Net & 19.82 & 0.81 & 0.34 & 16.64 & 0.59 & 0.53 & 14.98 & 0.72 & 0.36 \\
ViWS-Net & \textbf{19.91} & 0.80 & 0.33 & 18.60 & \textbf{0.70} & \textbf{0.41} & \textbf{15.90} & \textbf{0.80} & \underline{0.25} \\
\midrule
\rowcolor[rgb]{1.0,0.957,0.839}
\textbf{\ours{} (ours)} & 19.82 & \underline{0.81} & 0.33 & 16.63 & 0.59 & 0.53 & 14.98 & 0.73 & 0.36 \\
\bottomrule
\end{tabular}
\end{table*}

\section{Synthetic sand dust in Experiment 1}\label{sec:supp-dust}
\noindent\textbf{Setup.}
KITTI-C and ImageNet-C have no sand dust, so we synthesized it with a haze model whose airlight is sand-colored and whose transmission differs per color: $Y_c = X_c\,t_c + A_c(1-t_c)$, $t_c=\exp(-\beta_s k_c D)$, with $A=(0.80, 0.68, 0.48)$, $k=(0.55, 0.75, 1.0)$ for red, green and blue (blue is attenuated most), and a density map $D=0.5+0.5P$ from a fractal pattern $P$ drawn once per clip.
$\beta_s$ was fixed before the run so that the mean blue transmission is 0.75, 0.45 and 0.20 at severities 1, 3 and 5; the input PSNR at $k=1$ is \spDustNoneSA{}, \spDustNoneSB{} and \spDustNoneSC{}\,dB.
The scenes, settings and methods are those of \cref{sec:supp-weather}; the oracle InstructIR$^\ast$ receives ``Remove the dust from my photo'' (a task it was not trained for), the oracle Restormer$^\ast$ its blind weights.

\noindent\textbf{Results.}
The weather restorer ViWS-Net gains the most (\spDustViwsKoneGain{}\,dB at $k=1$, \spDustViwsKthirteenGain{}\,dB at $k=13$; \cref{tab:supp-dust}), followed by the oracle InstructIR$^\ast$ (\spDustInstrKoneGain{}\,dB), \ours{} (\spDustOursKoneGain{}\,dB at $k=1$, nothing at $k=13$) and AdaIR (\spDustAdaIRKoneGain{}\,dB), whereas PromptIR and MoCE-IR lose \spDustPromptIRKoneGain{} and \spDustMoCEIRKoneGain{}\,dB; the denoisers move the mean PSNR by at most \spDustDenoiserMaxAbsKone{}\,dB.
The weather check of \ours{} gains \spDustOursGainSA{}, \spDustOursGainSB{} and \spDustOursGainSC{}\,dB at severities 1, 3 and 5, is better in \spDustOursNbetter{} and worse in \spDustOursNworse{} of the \spDustNcases{} cases (worst \spDustOursWorst{}\,dB), and scales the three colors alike, so the yellow cast remains.

\begin{table}[t]
\centering
\caption{\textbf{Synthetic sand dust.} Exp.~1, Boreas, mean over \spDustNcases{} cases (6 scenes $\times$ severities 1, 3, 5). Ranking as in \cref{tab:supp-weather-k1}.}
\label{tab:supp-dust}
\scriptsize
\setlength{\tabcolsep}{2.2pt}
\begin{tabular}{@{}l ccc ccc@{}}
\toprule
 & \multicolumn{3}{c}{Current frame only ($k=1$)} & \multicolumn{3}{c}{All 13 frames ($k=13$)} \\
\cmidrule(lr){2-4}\cmidrule(lr){5-7}
Method & PSNR$\uparrow$ & SSIM$\uparrow$ & LPIPS$\downarrow$ & PSNR$\uparrow$ & SSIM$\uparrow$ & LPIPS$\downarrow$ \\
\midrule
No restoration & 18.41 & 0.86 & 0.19 & 18.41 & 0.86 & 0.19 \\
Median $3\times3$ & 18.24 & 0.80 & 0.27 & 18.24 & 0.80 & 0.27 \\
BM3D & 18.40 & 0.86 & 0.19 & 18.40 & 0.86 & 0.19 \\
\cmidrule(l{2pt}r{2pt}){1-7}
Restormer$^\ast$ & 18.41 & 0.86 & 0.19 & 18.41 & 0.86 & 0.19 \\
NAFNet & 17.97 & 0.79 & 0.34 & 17.97 & 0.79 & 0.34 \\
MT & 17.76 & 0.76 & 0.39 & 17.76 & 0.76 & 0.39 \\
SCUNet & 18.30 & 0.82 & 0.25 & 18.30 & 0.82 & 0.25 \\
DRUNet & 18.40 & 0.86 & 0.19 & 18.40 & 0.86 & 0.19 \\
FFDNet & 18.40 & 0.86 & 0.19 & 18.40 & 0.86 & 0.19 \\
\cmidrule(l{2pt}r{2pt}){1-7}
AdaIR & 20.26 & \underline{0.91} & \underline{0.18} & \underline{20.26} & \underline{0.91} & \underline{0.18} \\
PromptIR & 17.04 & 0.84 & 0.19 & 17.04 & 0.84 & 0.19 \\
MoCE-IR & 17.47 & 0.85 & 0.20 & 17.47 & 0.85 & 0.20 \\
InstructIR$^\ast$ & 21.38 & 0.93 & 0.17 & 21.38 & 0.93 & 0.17 \\
\cmidrule(l{2pt}r{2pt}){1-7}
FastDVDnet & 18.43 & 0.85 & 0.20 & 18.43 & 0.85 & 0.20 \\
RVRT & 18.40 & 0.86 & 0.19 & 18.40 & 0.86 & 0.19 \\
Shift-Net & 18.39 & 0.85 & 0.20 & 18.40 & 0.86 & 0.19 \\
ViWS-Net & \textbf{24.07} & \textbf{0.92} & \textbf{0.14} & \textbf{23.48} & \textbf{0.92} & \textbf{0.15} \\
\midrule
\rowcolor[rgb]{1.0,0.957,0.839}
\textbf{\ours{} (ours)} & \underline{20.49} & 0.90 & 0.19 & 18.41 & 0.86 & 0.19 \\
\bottomrule
\end{tabular}
\end{table}

\section{Rain, snow and fog in Experiment 2}\label{sec:supp-exp2-weather}
In Exp.~2 (Sec.~\ref{sec:protocol-kitti}) every frame is degraded, so this is the $k=13$ setting.
The noise indicator of \ours{} labels every weather frame noise-free, and the weather check finds no past reference: it returns at least \spEtwoOursKeptMin{}\% of the weather frames unchanged and changes the agreement and mAP by at most \spEtwoOursMaxAbs{} points.
Restorers trained on haze or weather keep many more decisions and detections under fog (\cref{tab:supp-exp2-weather}): PromptIR keeps \spEtwoBestFogAgree{}\% of the decisions (no restoration \spEtwoNoneFogAgree{}\%), and MoCE-IR reaches mAP \spEtwoBestFogMap{}\% (no restoration \spEtwoNoneFogMap{}\%); under rain and snow PromptIR keeps the most decisions (\spEtwoBestRainAgree{} and \spEtwoBestSnowAgree{}\%) and ViWS-Net the most detections (\spEtwoBestRainMap{} and \spEtwoBestSnowMap{}\%).

\begin{table}[t]
\centering
\caption{\textbf{Exp.~2 under weather} (KITTI, severity 3, all frames degraded): decisions as on the clean frame and detection mAP@0.5 (\%), metrics as in the main text. Ranking as in \cref{tab:supp-weather-k1}.}
\label{tab:supp-exp2-weather}
\scriptsize
\setlength{\tabcolsep}{2.6pt}
\begin{tabular}{@{}l cc cc cc@{}}
\toprule
 & \multicolumn{2}{c}{Rain} & \multicolumn{2}{c}{Snow} & \multicolumn{2}{c}{Fog} \\
\cmidrule(lr){2-3}\cmidrule(lr){4-5}\cmidrule(lr){6-7}
Method & decisions & mAP & decisions & mAP & decisions & mAP \\
\midrule
No restoration & 59.18 & 67.58 & 60.30 & 50.30 & 35.87 & 70.49 \\
Median $3\times3$ & 55.78 & 47.02 & 61.97 & 28.91 & 36.12 & 49.21 \\
\cmidrule(l{2pt}r{2pt}){1-7}
Restormer$^\ast$ & 65.24 & 78.44 & 60.30 & 50.28 & 35.87 & 70.48 \\
NAFNet & 62.15 & 42.69 & 63.02 & 25.41 & 38.16 & 40.90 \\
MT & 55.47 & 36.32 & 65.37 & 25.95 & 37.35 & 14.05 \\
SCUNet & 66.23 & 51.19 & 61.60 & 37.42 & 42.42 & 35.22 \\
DRUNet & 60.98 & 67.36 & 60.92 & 49.15 & 36.92 & 70.38 \\
FFDNet & 60.17 & 66.35 & 60.79 & 47.96 & 36.49 & 69.53 \\
\cmidrule(l{2pt}r{2pt}){1-7}
AdaIR & 66.36 & 67.97 & \underline{68.09} & \underline{54.35} & 67.78 & 84.68 \\
PromptIR & \textbf{72.73} & 66.88 & \textbf{69.14} & 52.57 & \textbf{70.81} & \underline{85.63} \\
MoCE-IR & \underline{68.95} & \underline{68.61} & 68.03 & 53.55 & \underline{69.51} & \textbf{85.86} \\
InstructIR$^\ast$ & 62.40 & 74.61 & 64.01 & 60.36 & 62.46 & 82.72 \\
\cmidrule(l{2pt}r{2pt}){1-7}
FastDVDnet & 61.66 & 64.54 & 61.16 & 46.16 & 37.79 & 64.05 \\
RVRT & 60.05 & 66.89 & 60.92 & 48.65 & 36.55 & 69.69 \\
Shift-Net & 60.61 & 66.39 & 60.73 & 48.20 & 37.04 & 68.88 \\
ViWS-Net & 63.27 & \textbf{71.16} & 56.77 & \textbf{56.96} & 65.37 & 67.88 \\
\midrule
\rowcolor[rgb]{1.0,0.957,0.839}
\textbf{\ours{} (ours)} & 59.25 & 67.55 & 60.24 & 50.33 & 35.81 & 70.54 \\
\bottomrule
\end{tabular}
\end{table}

\section{Gaps per condition in Experiment 1}\label{sec:supp-gaps}
\begin{figure*}[t]
\centering
\includegraphics[width=\textwidth]{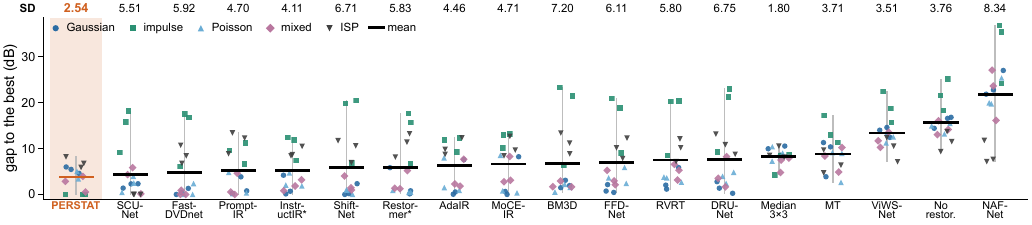}
\caption{\textbf{The gap of every method to the best one in each condition} (Exp.~1, Boreas, $k=1$). PSNR gap to the best method in each of the \nvNcond{} conditions (noise type $\times$ severity), methods sorted by the mean gap (bar); above: SD of the gaps (dB). Every learned method falls ${\geq}$\nvLearnedImpWorstMinKone{}\,dB behind under some impulse condition, \ours{} at most \nvOursWorstKone{}\,dB anywhere.}
\label{fig:supp-gaps}
\end{figure*}

\Cref{fig:supp-gaps} enlarges Fig.~\ref{fig:balance}b, with the methods as columns and the SD of the gaps of each method above; the points of Fig.~\ref{fig:balance}a are the mean and the SD of each column here.

\noindent\textbf{Leaving out one noise type.}
\Cref{tab:supp-loto} recomputes the worst-case and mean gaps of Exp.~1 without the three conditions of one noise type.
Without Gaussian, Poisson, mixed or ISP noise, \ours{} keeps the smallest worst-case and mean gaps at $k=1$ and $k=13$.
Without impulse noise, SCUNet has the smallest worst-case gap (\spLotoImpScunetWorst{} \vs \spLotoImpOursWorstKone{}\,dB for \ours{} at $k=1$ and \spLotoImpOursWorstKthirteen{}\,dB at $k=13$), and \ours{} ranks \spLotoImpOursMeanRankKone{} ($k=1$) and \spLotoImpOursMeanRankKthirteen{} ($k=13$) by the mean gap.
The lead of \ours{} therefore comes from impulse noise, under which every learned method falls ${\geq}$\nvLearnedImpWorstMinKone{}\,dB behind (\cref{fig:supp-gaps}); under the other four noise types, \ours{} is not the best, but its worst-case gap stays the second smallest.

\begin{table*}[t]
\centering
\caption{\textbf{Leaving out one noise type} (Exp.~1, Boreas). Worst-case and mean PSNR gaps (dB) to the best method over the 12 conditions that remain (all 15 for None), with the rank of \ours{} and the best other method. As in the main text, the methods told the noise type ($^\ast$) are not ranked. Bold: the smallest gap.}
\label{tab:supp-loto}
\small
\setlength{\tabcolsep}{4pt}
\begin{tabular}{@{}l cc cc cc cc@{}}
\toprule
 & \multicolumn{4}{c}{Current frame only ($k=1$)} & \multicolumn{4}{c}{All 13 frames ($k=13$)} \\
\cmidrule(lr){2-5}\cmidrule(lr){6-9}
 & \multicolumn{2}{c}{Worst-case gap} & \multicolumn{2}{c}{Mean gap} & \multicolumn{2}{c}{Worst-case gap} & \multicolumn{2}{c}{Mean gap} \\
\cmidrule(lr){2-3}\cmidrule(lr){4-5}\cmidrule(lr){6-7}\cmidrule(lr){8-9}
Left out & \ours{} & Best other & \ours{} & Best other & \ours{} & Best other & \ours{} & Best other \\
\midrule
None & \textbf{8.18} (1st) & 10.54 (Median) & \textbf{3.78} (1st) & 4.42 (SCUNet) & \textbf{7.90} (1st) & 10.53 (Median) & \textbf{3.03} (1st) & 4.12 (SCUNet) \\
\midrule
Gaussian & \textbf{8.18} (1st) & 10.53 (Median) & \textbf{3.39} (1st) & 5.02 (SCUNet) & \textbf{7.90} (1st) & 10.53 (Median) & \textbf{2.85} (1st) & 4.83 (SCUNet) \\
Impulse & 8.18 (2nd) & \textbf{5.83} (SCUNet) & 4.72 (7th) & \textbf{1.93} (SCUNet) & 7.90 (2nd) & \textbf{5.83} (SCUNet) & 3.78 (3rd) & \textbf{1.60} (SCUNet) \\
Poisson & \textbf{8.18} (1st) & 10.54 (Median) & \textbf{3.68} (1st) & 4.95 (SCUNet) & \textbf{7.90} (1st) & 10.53 (Median) & \textbf{3.27} (1st) & 4.66 (SCUNet) \\
Mixed & \textbf{8.18} (1st) & 10.54 (Median) & \textbf{4.11} (1st) & 4.66 (SCUNet) & \textbf{7.90} (1st) & 10.53 (Median) & \textbf{3.18} (1st) & 4.35 (SCUNet) \\
ISP & \textbf{5.99} (1st) & 10.54 (Median) & \textbf{2.99} (1st) & 3.56 (PromptIR) & \textbf{4.36} (1st) & 9.95 (Median) & \textbf{2.05} (1st) & 3.19 (PromptIR) \\
\bottomrule
\end{tabular}
\end{table*}

\section{Constants of \ours{}}\label{sec:supp-constants}
\Cref{tab:supp-constants} lists every constant of \ours{} and how we set it.
We use the KITTI validation drives 0091, 0095 and 0096, which appear in neither experiment, as a held-out validation set.
On this set, before the runs of Exp.~1 and 2, we selected four choices by comparing alternatives (val.): the outlier threshold of the card ($6\sigma_r$ against $4\sigma_r$), the saturated path of the outlier test (with against without), the impulse threshold (0.5\% against 0.2\%) and the unflag rule of color completion (unflagging only, against a rule that also decides from the chroma which channels are corrupted).
We set the other constants by hand without comparing values (hand), and some follow from theory (theory).
Two of them depend on the camera and the use rather than on the noise: the window length (13 frames at 5\,Hz) and the number of reference frames of the weather check (3).
We chose these values for this paper, and they can be adapted to the frame rate and the application in practice.
The thresholds of the noise indicator are measured as 1.25 times the largest score on clean frames of the same camera (KITTI validation drives for KITTI; for Boreas, 1{,}756 frames of three clear drives).
One of these Boreas drives is the drive of Exp.~1, but recomputing the thresholds without it (1{,}200 frames of the other two drives) gives identical values, because both maxima lie on another drive, so no label of Exp.~1 changes (0 of 4{,}290 frames).
Apart from the window (below), we have not measured the sensitivity of the results to these constants.

\begin{table*}[t]
\centering
\caption{\textbf{Constants of \ours{}} and how they were set. val.: selected on the held-out KITTI validation drives by comparing alternatives, before the runs; hand: set by hand without comparing values; meas.: measured on clean frames; theory: follows from the stated reason.}
\label{tab:supp-constants}
\small
\setlength{\tabcolsep}{4pt}
\begin{tabular}{@{}llll@{}}
\toprule
Step & Constant & Value & How set \\
\midrule
Frame tensor & frames in the window & 13, \ie 12 past & hand (adaptable) \\
Noise indicator & noise score & 10\% quantile over $16\times16$ blocks & hand \\
 & isolated outlier & ${>}0.25$ from the $3\times3$ median, ${>}0.125$ from neighbors & hand \\
 & thresholds $\tau_\sigma,\tau_{\rm imp}$ & $1.25\times$ the largest value on clean frames & meas.; 1.25 hand \\
 & MAD factor & 1.4826 & theory (Gaussian SD) \\
Weather check & score threshold & 0.3 & hand \\
 & reference & 3 latest frames of the past reference & hand (adaptable) \\
 & window, gain range & $H/8$, gain clipped to $[0.25, 6]$ & hand \\
Noise card & grid & $4\times8$ regions & hand \\
 & past reference needed & $|\cC_t|\ge4$ & hand \\
 & outlier threshold & $|r|>\max(6\sigma_r, 0.10)$ & val. ($6$ \vs $4\sigma_r$); 0.10 hand \\
 & outlier shape & 7 of 8 neighbors on one side & hand \\
 & saturated path & saturated in all channels & val. (with \vs without) \\
 & impulse & excess ${\ge}0.5\%$ & val. (0.5\% \vs 0.2\%) \\
 & Gaussian & $\widehat\sigma_t\ge3/255$ & hand \\
Color completion & unflag rule & unflag if $z\le4$ & val. (rule); 4 hand \\
 & neighborhood, floor & $5\times5$, $\epsilon=2/255$ & hand \\
 & solvable system & $|\det|\ge0.05$, else missing & hand \\
Color step & tile, shifts & $16\times16$, shifts 0 and 8 & hand \\
 & $\lambda_t$ & $1.5\,\widehat\sigma_t\sqrt{255}$ & 1.5 hand; $\sqrt{255}$ theory \\
 & $\lambda_t$, no Gaussian component & RMS of $D(o)$ in place of $\widehat\sigma_t$ & hand \\
TV & iterations & 300 & hand \\
 & step & $0.99/\sqrt8$ & theory (convergence) \\
 & $\delta_t$, no Gaussian component & $\delta_t^2=3|\Omega|q^2$, $q=1/(255\sqrt{12})$ & theory (8-bit quantization) \\
\bottomrule
\end{tabular}
\end{table*}

\noindent\textbf{Window length and noisy frames.}
We use a window of 13 frames in this paper.
\Cref{tab:supp-window} reruns \ours{} on every case of Exp.~1 with windows of 5 to 21 frames, and, for 13 frames, with the last $k$ frames degraded; the other methods keep their rows of Exp.~1, which exist only for $k=1$ and $k=13$.
The window length barely matters: the worst-case gap to the best method changes by at most 0.05\,dB, it stays the smallest of all methods for every length, the weather and clean results do not change, and only the time grows (277.78 to 403.01\,ms per frame).
The window can therefore be set by the frame rate and the latency budget of the vehicle.
With $k$ noisy frames, the card compares with the past reference while at least four clean frames remain ($k\le9$) and otherwise estimates the noise without it, and the worst-case gap stays between 7.79 and 8.46\,dB, the smallest of all methods against their rows at $k=1$.
The weather check, in contrast, helps only during the first frames of weather (Sec.~\ref{sec:supp-details}).

\begin{table}[t]
\centering
\caption{\textbf{Sensitivity to the window} (Exp.~1, \ours{} rerun; gaps in dB to the best method over the 15 noise conditions, the other methods from Exp.~1). Top, window length $W$ at the onset ($k=1$) and with all $W$ frames degraded ($k=W$, gaps against the $k=13$ rows), and the median time per frame. Bottom, $W=13$ with the last $k$ frames degraded (gaps against the $k=1$ rows) and the PSNR gain of the weather check on rain, snow and fog.}
\label{tab:supp-window}
\small
\setlength{\tabcolsep}{2.4pt}
\begin{tabular}{@{}rccccr@{}}
\toprule
 & \multicolumn{2}{c}{$k=1$} & \multicolumn{2}{c}{$k=W$} & \\
$W$ & worst & mean & worst & mean & ms \\
\midrule
5 & 8.17 & 3.78 & 7.90 & 3.06 & 277.78 \\
9 & 8.17 & 3.78 & 7.90 & 3.05 & 313.67 \\
13 & 8.18 & 3.78 & 7.90 & 3.03 & 339.16 \\
17 & 8.18 & 3.78 & 7.93 & 3.02 & 371.11 \\
21 & 8.18 & 3.78 & 7.95 & 3.04 & 403.01 \\
\bottomrule
\end{tabular}

\medskip
\begin{tabular}{@{}rrccc@{}}
\toprule
$k$ & clean past & worst & mean & weather gain \\
\midrule
1 & 12 & 8.18 & 3.78 & 6.07/4.24/8.55 \\
3 & 10 & 8.25 & 3.79 & 2.48/1.86/3.08 \\
5 & 8 & 8.33 & 3.80 & 0.00/0.00/0.00 \\
7 & 6 & 8.41 & 3.81 & 0.00/0.00/0.00 \\
9 & 4 & 8.46 & 3.84 & 0.00/0.00/0.00 \\
10 & 3 & 7.79 & 3.28 & 0.00/0.00/0.00 \\
11 & 2 & 7.83 & 3.28 & 0.00/0.00/0.00 \\
13 & 0 & 7.90 & 3.29 & 0.00/0.00/0.00 \\
\bottomrule
\end{tabular}
\end{table}

\section{Details of the method and the protocol}\label{sec:supp-details}
\noindent\textbf{Pseudocode.}
\Cref{alg:ours} gives \ours{} for one frame.

\begin{algorithm}[t]
\caption{\ours{} for one frame (Steps 1--4 of Fig.~\ref{fig:overview} and Sec.~\ref{sec:steps}).}\label{alg:ours}
\small
\begin{algorithmic}[1]
\Require frame tensor $\cY_t=[Y_{t-12},\dots,Y_t]$
\State $\cC_t\gets\{s<t:g(Y_s)=0\}$\Comment{(1) Eq.~\eqref{eq:gate}}
\If{$g(Y_t)=0$}\Comment{weather check}
  \State \Return $Y_t$, or $Y_t$ matched to $\mathcal{R}_t\subset\cC_t$ if weather-degraded
\EndIf
\State $\widehat\sigma_t,\widehat p_t\gets$ card of $Y_t$ against $\cC_t$\Comment{(2) Eq.~\eqref{eq:card}}
\State $b_t\gets Y_t$;\ \ $m_{ij}\gets1$ for all $i,j$
\If{$\widehat p_t$ indicates impulses}\Comment{(3) color completion}
  \State $Q\gets$ eigenvectors of $\unf{\cP}{4}\unf{\cP}{4}^\top$\Comment{Eq.~\eqref{eq:factor}}
  \State flag channels per color; unflag pixels that fit\Comment{Eq.~\eqref{eq:supp-veto}}
  \ForAll{pixels $y$ of $b_t$ with flagged channels $B\neq\emptyset$}
    \If{$|B|\le2$} complete $y_B$ by \eqref{eq:fiber}
    \Else\ $m_{ij}\gets0$;\ \ $y\gets$ its $3\times3$ median
    \EndIf
  \EndFor
  \State $\widehat\sigma_t\gets$ \eqref{eq:card} on unflagged $2\times2$ blocks
\EndIf
\State $U,w\gets$ color axes of the unfiltered past\Comment{(4) Eq.~\eqref{eq:split}}
\State $b'_t\gets b_tUU^\top+o'w^\top$, $o'$ by \eqref{eq:shrink}\Comment{closed form}
\State \Return $\widehat V_t\gets$ solution of \eqref{eq:tv} from $b'_t$, clipped
\end{algorithmic}
\end{algorithm}

\noindent\textbf{Outlier test of the noise card.}
A pixel is an outlier in a color channel if it lies far from its $3\times3$ median with almost all of its neighbors on one side, or if it is saturated in all channels; $\rho(Y)$ is the share of such pixels.
With fewer than four frames in the past reference, the card uses $\widehat p_t=\rho(Y_t)$, and $\widehat\sigma_t$ is the MAD estimate on $D$ of the noisy frames.

\noindent\textbf{Unflag rule of color completion.}
The outlier test also flags wires and window frames, but these change the three colors together, so their chroma fits the neighborhood.
A pixel with one or two flagged channels, none saturated, is therefore kept as a real structure (unflagged) if its chroma lies within four spreads of the neighborhood,
\begin{equation}\label{eq:supp-veto}
z(y)=\bigl\|\bigl((c_a(y)-\tilde c_a)/s_a\bigr)_{a=2,3}\bigr\|_2\le4 ,
\end{equation}
with $s_a^2=(1.4826\MAD_a)^2+\epsilon^2$ and $c_a$, $\tilde c_a$, $\MAD_a$, $\epsilon$ as in Sec.~\ref{sec:steps}.
The bound 4 accepts up to four times the spread; it is a hyperparameter and can be adapted in practice.
After the completion, $\widehat\sigma_t$ is measured again by Eq.~\eqref{eq:card} on the $2\times2$ blocks without flagged channels.

\noindent\textbf{Two tilings of the color step.}
The color step shrinks $o$ on two tilings of $16\times16$ tiles, shifted by 8 pixels, and averages the two results; $\bar\gamma_{ij}$ is the mean of the two shrink factors at pixel $(i,j)$.
Without a Gaussian component, the RMS of the Haar diagonal detail $D(o)$ replaces $\widehat\sigma_t$ in $\lambda_t$.

\noindent\textbf{Weather check.}
Let $L(Y)$ be the mean of R, G and B at each pixel, and $\kappa(Y)=\operatorname{mean}_{ij}\bigl(\max_cY_{ijc}-\min_cY_{ijc}\bigr)$ its colorfulness.
With bars denoting means over the three latest frames of the past reference $\mathcal{R}_t$, the weather check returns the frame unchanged if
\begin{equation}\label{eq:supp-weather}
\Bigl|\ln\frac{\mathrm{std}\,L(Y_t)}{\overline{\mathrm{std}\,L}}\Bigr|+\Bigl|\ln\frac{\kappa(Y_t)}{\bar\kappa}\Bigr|<0.3 .
\end{equation}
Otherwise, a gain field and an offset field, shared by all colors, match the local mean and standard deviation of $L$ (window $H/8$, gain clipped to $[0.25,6]$) to those of $\mathcal{R}_t$.
The TV of Step 4 (Eq.~\eqref{eq:tv}) then fills thin bright structures in excess of those of $\mathcal{R}_t$ (streaks, flakes).
In this paper, $\mathcal{R}_t$ holds the three latest frames of the past reference, which keeps the reference close in time to $Y_t$ and still averages over more than one frame; this number can be adapted in practice (Sec.~\ref{sec:supp-constants}).
Because the noise indicator tests only noise, weather-degraded past frames enter $\mathcal{R}_t$, so the weather check can only undo weather during its first frames: it raises rain, snow and fog by 6.07, 4.24 and 8.55\,dB when only $Y_t$ is degraded, by 2.48, 1.86 and 3.08\,dB when the last three frames are, and not at all from five frames on (\cref{tab:supp-window}).
Its gain is the same for all colors, so a color cast remains.

\noindent\textbf{Where the past enters.}
The past enters across time through the noise card and the weather check, and in the color space through the principal color axes $Q$ and $U$.
Noise that is independent and equal across colors leaves these eigenvectors unchanged in expectation.

\noindent\textbf{Noise parameters.}
Gaussian noise has $\sigma=20.4$, 45.9 and 96.9 on the 8-bit scale at severities 1, 3 and 5.
Impulse noise sets each color channel of each pixel to 0 or 255 with probabilities 3, 9 and 27\%.
Poisson noise is $\mathrm{Poisson}(cx)/c$ with $c=\nvShotA,\nvShotB,\nvShotC$.
ISP noise adds shot and read noise in raw space after inverting the camera pipeline, and only its difference after the pipeline is added to the clean frame, so on mid-gray it is as strong as the Gaussian noise.

\noindent\textbf{Compared methods.}
Restormer$^\ast$ is given the degradation type and uses the matching non-blind Gaussian weights ($\sigma=15,25,50$), deraining weights or blind weights.
NAFNet uses blind weights for real noise (SIDD), MT is trained on Gaussian noise with $\sigma=15$, and SCUNet is blind and trained on synthetic Gaussian, Poisson, speckle, JPEG and camera noise.
AdaIR, PromptIR and MoCE-IR are blind, with one model for Gaussian noise ($\sigma=15$--50), rain and haze.
InstructIR$^\ast$ covers seven tasks and receives one fixed instruction per degradation type.
FastDVDnet and ViWS-Net run on the frames $(t-2,t-1,t,t-1,t-2)$, and RVRT and Shift-Net (both non-blind, $\sigma\in[0,50]$) run on $t-12,\dots,t$ with their last output kept.
FastDVDnet, RVRT and Shift-Net receive noise maps from $\widehat\sigma$, the wavelet MAD estimate averaged over the colors.

\noindent\textbf{Exp.~2: start of a drive and paired intervals.}
Before the 13th frame of a drive, the window repeats the first frame of the drive.
\Cref{tab:supp-paired} gives the paired differences of Sec.~\ref{sec:exp-downstream} with their 95\% intervals.
We resample 10-frame blocks of each drive (2{,}000 resamples, the same for both methods), recompute the agreement and the mAP of the resampled frames, and take the 2.5 and 97.5\% quantiles of the difference.
For the worse noise type, each resample takes the lower agreement of Gaussian and impulse noise for each method.
Averaged over the two noise types, every interval excludes zero, but under the worse noise type the intervals against SCUNet, PromptIR and InstructIR$^\ast$ include zero (italics).
The intervals cover these \npNframes{} frames of six drives, not other drives.

\begin{figure*}[t]
\centering
\input{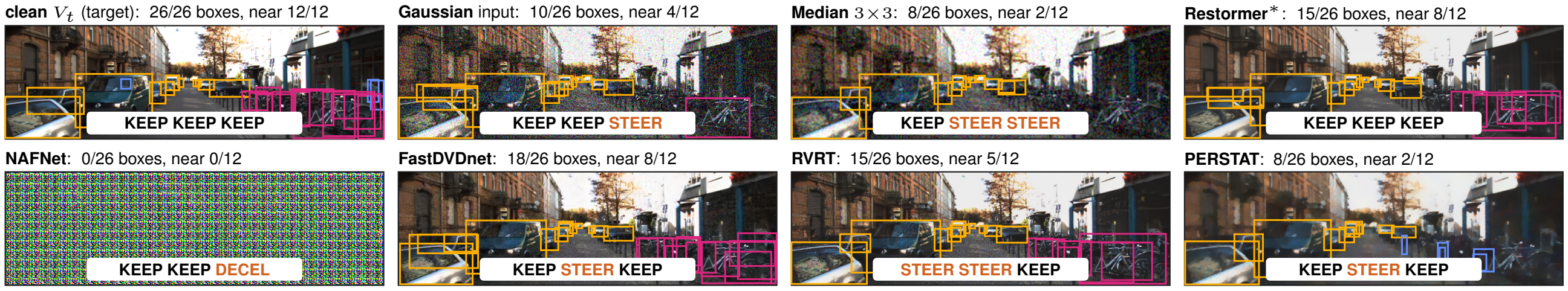}
\caption{\textbf{Exp.~2, the frame of Fig.~\ref{fig:exp2-design}b under Gaussian noise} (severity 3). Boxes, counts and decision chips as in Fig.~\ref{fig:exp2-design}b: $n/26$ boxes of the clean $V_t$ found again, the decisions of the 3 seeds (orange: differs from $V_t$). Under Gaussian noise, the oracle Restormer$^\ast$ gives the same image as Restormer with estimated noise.}
\label{fig:supp-gauss}
\end{figure*}

\noindent\textbf{Exp.~2: the example frame under Gaussian noise.}
\Cref{fig:supp-gauss} shows the frame of Fig.~\ref{fig:exp2-design}b under Gaussian noise for every method with example frames.
\ours{} finds 8 of the 26 boxes and 2 of the 12 near ones and loses the bicycles, whereas FastDVDnet finds 18 and the oracle Restormer$^\ast$ and RVRT 15.
Methods added after the example run, such as Shift-Net, the best other method under Gaussian noise in Tab.~\ref{tab:onset-downstream}, have no example frames.
NAFNet finds no box under either noise type, as its mAP of 0.00\% in Tab.~\ref{tab:onset-downstream} shows.

\begin{table}[t]
\centering
\caption{\textbf{Paired differences in Exp.~2} (\ours{} minus each method, percentage points, with 95\% intervals). Noise: Gaussian and impulse noise together (decisions averaged, mAP on the pooled frames). Worse noise type: the lower agreement of the two types for each method. Italics: the interval includes zero.}
\label{tab:supp-paired}
\scriptsize
\setlength{\tabcolsep}{2.5pt}
\begin{tabular}{@{}l ccc@{}}
\toprule
 & \multicolumn{2}{c}{Noise (both types)} & Worse noise type \\
\cmidrule(lr){2-3}\cmidrule(lr){4-4}
\ours{} minus & Decisions (points) & mAP (points) & Decisions (points) \\
\midrule
No restoration & 30.71 [26.27, 35.31] & 31.54 [29.14, 33.30] & 22.70 [17.47, 27.77] \\
Median $3\times3$ & 16.67 [13.52, 19.69] & 10.48 [8.06, 13.34] & 21.77 [17.32, 25.99] \\
\cmidrule(l{2pt}r{2pt}){1-4}
Restormer$^\ast$ & 13.08 [8.69, 17.81] & 19.64 [16.08, 22.56] & 18.74 [10.88, 26.78] \\
NAFNet & 35.10 [28.07, 42.49] & 61.35 [58.85, 63.56] & 23.50 [14.94, 32.65] \\
MT & 30.92 [26.31, 35.74] & 24.43 [21.63, 26.74] & 24.49 [19.48, 29.44] \\
SCUNet & 4.64 [1.55, 7.82] & 17.35 [13.83, 20.44] & \textit{2.41 [-2.84, 7.43]} \\
DRUNet & 17.19 [13.33, 21.17] & 24.18 [21.55, 26.43] & 20.35 [13.89, 26.91] \\
FFDNet & 14.66 [11.50, 18.15] & 21.06 [18.13, 23.73] & 19.05 [12.63, 25.74] \\
\cmidrule(l{2pt}r{2pt}){1-4}
AdaIR & 11.26 [6.08, 16.73] & 20.61 [17.47, 23.44] & 13.48 [5.44, 21.83] \\
PromptIR & 6.52 [2.19, 10.85] & 20.51 [17.26, 23.26] & \textit{3.96 [-2.41, 10.62]} \\
MoCE-IR & 13.82 [9.01, 18.86] & 17.02 [13.78, 20.29] & 19.05 [11.21, 26.98] \\
InstructIR$^\ast$ & 5.81 [0.96, 10.99] & 15.12 [11.75, 18.69] & \textit{3.77 [-3.48, 11.54]} \\
\cmidrule(l{2pt}r{2pt}){1-4}
FastDVDnet & 23.35 [20.35, 26.38] & 18.12 [14.70, 21.68] & 29.19 [24.24, 34.08] \\
RVRT & 18.55 [15.02, 22.40] & 17.90 [14.57, 21.59] & 23.56 [17.50, 29.87] \\
Shift-Net & 16.82 [13.23, 20.66] & 16.40 [12.82, 19.98] & 26.04 [20.05, 32.22] \\
ViWS-Net & 33.58 [28.90, 38.10] & 30.28 [27.98, 32.14] & 23.38 [18.18, 28.51] \\
\bottomrule
\end{tabular}
\end{table}

\end{document}